\pdfoutput=1
\documentclass{article}

\PassOptionsToPackage{numbers, sort&compress}{natbib}

\usepackage[main, final]{neurips_2026}

\usepackage[utf8]{inputenc} 
\usepackage[T1]{fontenc}    
\usepackage{hyperref}       
\usepackage{url}            
\usepackage{booktabs}       
\usepackage{multirow}       
\usepackage{colortbl}       
\usepackage{graphicx}       
\usepackage{afterpage}      
\usepackage{wrapfig}        
\usepackage{subcaption}     
\usepackage{array}          
\usepackage{amsfonts}       
\usepackage{nicefrac}       
\usepackage{microtype}      
\usepackage{xcolor}         

\hypersetup{
  pdftitle={NS-VLA: Towards Neuro-Symbolic Vision-Language-Action Models},
  pdfauthor={Ziyue Zhu, Shangyang Wu, Shuai Zhao, Zhiqiu Zhao, Jian Zhang, Shengjie Li, Yi Wang, Anh Tuan Luu, Xinliang Zhou, Fang Li, and Haoran Luo},
  pdfsubject={NeurIPS 2026},
  pdfkeywords={vision-language-action, neuro-symbolic learning, robotic manipulation}
}

\usepackage{amsmath}
\usepackage{amssymb}
\usepackage{mathtools}
\usepackage{amsthm}

\usepackage{algorithm}
\usepackage{algpseudocode}
\algrenewcommand\alglinenumber[1]{\footnotesize #1:}

\usepackage[capitalize,noabbrev]{cleveref}

\usepackage[most]{tcolorbox}
\usepackage{listings}

\DeclareMathOperator*{\argmax}{argmax}
\DeclareMathOperator*{\argmin}{argmin}

\definecolor{promptbg}{RGB}{252,235,221}
\definecolor{promptbd}{RGB}{40,40,40}

\lstdefinestyle{promptstyle}{
  basicstyle=\ttfamily\small,
  columns=fullflexible,
  breaklines=true,
  keepspaces=true,
  showstringspaces=false
}

\newtcblisting{promptbox}{
  enhanced,
  colback=promptbg,
  colframe=promptbd,
  boxrule=0.8pt,
  arc=1.2mm,
  left=6pt,right=6pt,top=6pt,bottom=6pt,
  drop shadow,
  listing only,
  listing options={style=promptstyle},
  width=0.96\linewidth
}

\definecolor{exbg}{RGB}{230,243,255}
\definecolor{exbd}{RGB}{40,40,40}

\newtcblisting{examplebox}{
  enhanced,
  colback=exbg,
  colframe=exbd,
  boxrule=0.8pt,
  arc=1.2mm,
  left=6pt,right=6pt,top=10pt,bottom=6pt,
  drop shadow,
  listing only,
  listing options={style=promptstyle},
  width=0.96\linewidth,
  title=Example,
  fonttitle=\bfseries\ttfamily\small,
  coltitle=black,
  attach boxed title to top left={xshift=6pt,yshift=-2mm},
  boxed title style={
    colback=exbg,
    colframe=exbd,
    boxrule=0.8pt,
    arc=1mm,
    left=4pt,right=4pt,top=2pt,bottom=2pt
  }
}

\definecolor{nsboxfill}{HTML}{F3F9FE}
\definecolor{nsboxframe}{HTML}{1597D8}
\newtcblisting{nsbox}[2]{
  enhanced, colback=nsboxfill, colframe=nsboxframe, boxrule=0.7pt, arc=1.2mm,
  left=5pt, right=5pt, top=3pt, bottom=3pt, drop fuzzy shadow,
  listing only, listing options={style=promptstyle, basicstyle=\ttfamily\footnotesize},
  title={#1}, fonttitle=\bfseries\small, coltitle=white, colbacktitle=nsboxframe,
  toptitle=1pt, bottomtitle=1pt, equal height group=#2
}

\theoremstyle{plain}
\newtheorem{theorem}{Theorem}[section]
\newtheorem{proposition}[theorem]{Proposition}
\newtheorem{lemma}[theorem]{Lemma}

\theoremstyle{definition}

\theoremstyle{remark}

\newcommand{\se}[1]{\textsubscript{\fontsize{5pt}{6pt}\selectfont\color{black!75}\textpm#1}}
\newcommand{\colhead}[1]{{\fontsize{8pt}{10pt}\selectfont #1}}
\definecolor{gainpos}{HTML}{5C9C5A}
\newcommand{\sdelta}[1]{\textbf{\textcolor{gainpos!80!black}{#1}}}
\definecolor{lossneg}{HTML}{C0504D}
\newcommand{\sdrop}[1]{\textsubscript{\fontsize{6pt}{7pt}\selectfont\color{lossneg}\,\textminus#1}}
\definecolor{nsblue}{HTML}{2B7BD6}
\definecolor{nslight}{HTML}{EAF3FC}
\makeatletter
\newcommand{\shadeL}[1][0pt]{\gdef\CT@cell@color{\@tempdimb#1\relax\global\let\CT@cell@color\relax}}
\newcommand{\shadeR}[1][0pt]{\gdef\CT@cell@color{\@tempdimc#1\relax\global\let\CT@cell@color\relax}}
\newcommand{\cellcolorR}[1]{\gdef\CT@cell@color{\CT@color{#1}\@tempdimc\z@\global\let\CT@cell@color\relax}}
\makeatother
\definecolor{famimi}{HTML}{999999}\definecolor{famvla}{HTML}{C1653C}\definecolor{famns}{HTML}{3D6F9A}
\newcommand{\yes}{\textcolor{nsblue}{\ensuremath{\checkmark}}}
\newcommand{\no}{\textcolor{black!30}{\ensuremath{\times}}}

\title{NS-VLA: Towards Neuro-Symbolic Vision-Language-Action Models}

\author{%
  \textbf{Ziyue Zhu}{\normalfont\textsuperscript{1,}\thanks{Equal contribution.\qquad \textsuperscript{\textdagger}Corresponding authors.\qquad \textsuperscript{1}\url{https://github.com/Zuzuzzy/NS-VLA}}}
  \enspace \textbf{Shangyang Wu}{\normalfont\textsuperscript{1,}\footnotemark[1]}
  \enspace \textbf{Shuai Zhao}{\normalfont\textsuperscript{1,\textdagger}}
  \enspace \textbf{Zhiqiu Zhao}{\normalfont\textsuperscript{1}}
  \enspace \textbf{Jian Zhang}{\normalfont\textsuperscript{2}}
  \enspace \textbf{Shengjie Li}{\normalfont\textsuperscript{1}} \\
  \textbf{Yi Wang}{\normalfont\textsuperscript{2}}
  \enspace \textbf{Anh Tuan Luu}{\normalfont\textsuperscript{2,3}}
  \enspace \textbf{Xinliang Zhou}{\normalfont\textsuperscript{2}}
  \enspace \textbf{Fang Li}{\normalfont\textsuperscript{2}}
  \enspace \textbf{Haoran Luo}{\normalfont\textsuperscript{2,\textdagger}} \\
  \textsuperscript{1}Beijing University of Posts and Telecommunications \\
  \textsuperscript{2}Nanyang Technological University, Singapore \quad \textsuperscript{3}VinUniversity \\
  \texttt{\{zyzhu, wushangyang, zhaoshuaiby\}@bupt.edu.cn}, \texttt{haoran.luo@ntu.edu.sg}
}

\begin{document}

\maketitle
\setcounter{footnote}{0}

\begin{abstract}
Vision-Language-Action (VLA) models ground natural-language instructions in visual observations to generate actions for robotic manipulation. On long-horizon tasks, existing VLA models face structure-blind backbones, backbone-bound generalization, and flat single-objective optimization. To address these challenges, we propose a novel \textbf{N}euro-\textbf{S}ymbolic \textbf{V}ision-\textbf{L}anguage-\textbf{A}ction (NS-VLA) framework. It introduces a Neuro-Symbolic Encoder that parses the instruction into a primitive plan tracked by a gated monotone stage pointer, a Neuro-Symbolic Solver that conditions a chunked VLA policy on the active sub-instruction, and Hierarchical Joint Policy Optimization, which fits the pointer on demonstration segments and refines the action policy with success-filtered online learning. Even when trained on few demonstrations, NS-VLA clearly leads on LIBERO and the held-out LIBERO-Plus ($76.5\%$ and $51.8\%$ against $71.2\%$ and $45.4\%$ for its backbone), exceeds the strongest baseline by $12.7$ points on long-horizon tasks, and raises real-robot success from $40.0\%$ to $60.0\%$. Our code is publicly available.\footnotemark
\end{abstract}

\section{Introduction}
\label{Section1}

\begin{wrapfigure}{r}{0.5\textwidth}
  \vspace{-28pt}
  \centering
  \includegraphics[width=\linewidth]{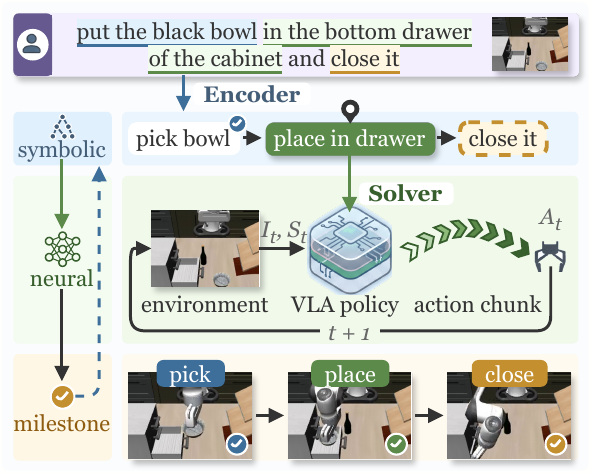}
  \vspace{-14pt}
  \caption{NS-VLA plans, then acts step by step.}
  \label{fig:teaser}
  \vspace{-10pt}
\end{wrapfigure}
Vision-Language-Action (VLA) models~\citep{RT-1, RT-2, Palm-E, Gato} map a natural-language instruction and visual observations to low-level robot actions (Figure~\ref{fig:teaser}) and are the main route toward generalist robots that follow free-form instructions~\citep{Open-X-Embodiment, Octo}. Many everyday instructions are long-horizon: ``put the black bowl in the bottom drawer of the cabinet and close it'' chains primitives that must each succeed in order. Such tasks are costly to demonstrate, so a practical VLA must learn them from a few examples.

\begin{figure*}[t]
\centering
\includegraphics[width=\linewidth]{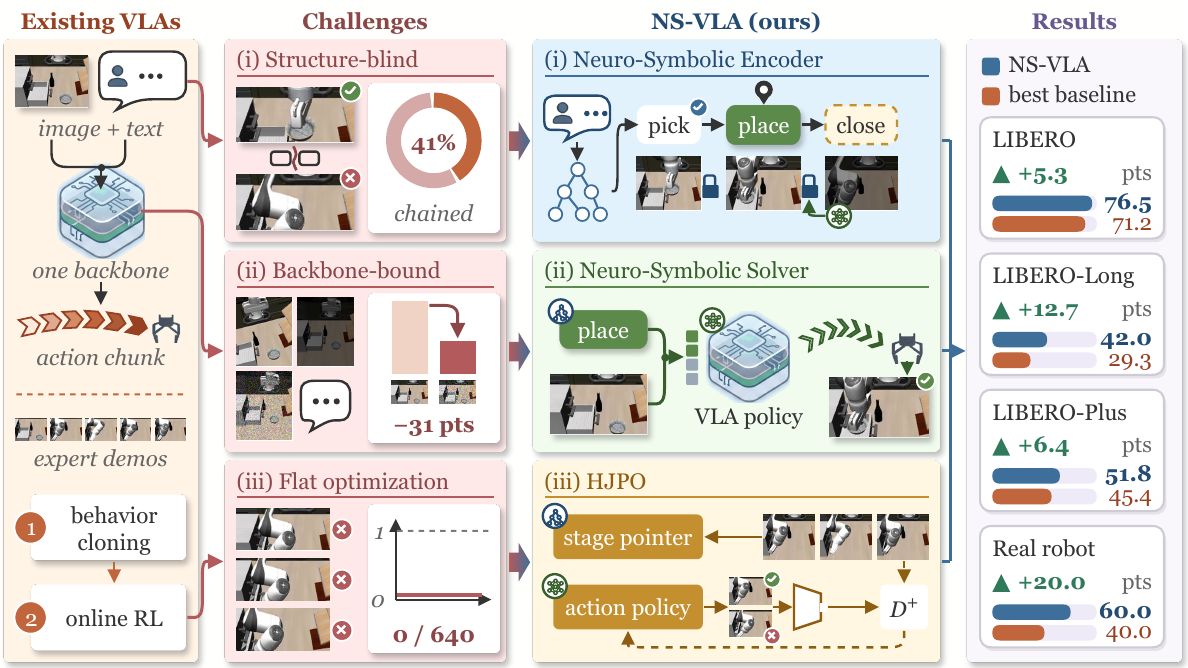}
\caption{Overview: existing VLAs face three challenges; NS-VLA answers each with its Encoder, Solver and HJPO, and clearly leads on LIBERO, LIBERO-Plus and a real robot by up to 20 points.}
\vspace{-10pt}
\label{f2}
\end{figure*}

Existing VLAs adopt a monolithic design: one backbone fuses vision and language, and a continuous action head decodes actions through diffusion~\citep{Diffusion-policy, 3D-VLA}, flow matching~\citep{PI_0, pi_0.5} or chunked regression~\citep{OpenVLA-OFT, VLA-Adapter}. Others fine-tune multimodal LLMs~\citep{RT-H, CoT-VLA, OpenVLA, ECoT, tracevla} or compact backbones~\citep{SmolVLA} to predict tokenized or chunked actions. Training pairs behavior cloning on expert demonstrations~\citep{finn2017one, Implicit-Behavioral-Cloning, BC-Z, Advances-in-Neural-Information-Processing-Systems} with online RL beyond them~\citep{VLA-RL, SimpleVLA-RL}, assuming the backbone internalizes the task decomposition, compositional generalization and training signal of long-horizon tasks.

However, this assumption raises three challenges on long-horizon tasks. \textbf{(i) Structure-blind backbones.} Tasks share reusable primitives~\citep{pmlr, Code-as-Policies, VoxPoser, Inner-Monologue}, yet end-to-end VLAs imitate a task as one undivided trajectory and cannot reuse a primitive in a new chain. \textbf{(ii) Backbone-bound generalization.} Generalization rests on one backbone, so paraphrases, new viewpoints, lighting or sensor noise cut success~\citep{LIBERO-Plus} beyond what the demonstrations cover. \textbf{(iii) Flat optimization.} Online RL maximizes one trajectory-level return of a flat policy~\citep{VLA-RL, SimpleVLA-RL, SRPO}, ignoring the task hierarchy~\citep{Hierarchical-Deep-Reinforcement-Learning, Learning-Multi-Level-Hierarchies-with-Hindsight}; as long tasks rarely succeed early, the binary reward is silent and partial progress unrewarded: flat RL on VLA-Adapter succeeds in none of its $640$ online rollouts on the LIBERO-Long suite.

We therefore propose a \textbf{N}euro-\textbf{S}ymbolic \textbf{V}ision-\textbf{L}anguage-\textbf{A}ction (\textbf{NS-VLA}) framework (Figure~\ref{f2}), answering the three challenges one to one. \textbf{First}, the \textbf{Neuro-Symbolic Encoder} parses the instruction into an ordered plan of primitives like pick, place and close, and tracks it with a monotone stage pointer that advances only when a neural classifier and a symbolic gripper gate agree. \textbf{Moreover}, the \textbf{Neuro-Symbolic Solver} turns the active primitive into a sub-instruction for a chunked VLA policy, so each demonstration segment teaches a reusable skill in one canonical wording; on single-primitive tasks it equals the backbone. \textbf{Furthermore}, \textbf{Hierarchical Joint Policy Optimization (HJPO)} fits the pointer on demonstration segments and refines the action policy by advantage-weighted regression on successful rollouts and demonstrations, giving each level its own learning signal.

We evaluate NS-VLA on LIBERO~\citep{LIBERO}, the held-out LIBERO-Plus~\citep{LIBERO-Plus} and a real robot arm. Trained on only a few demonstrations, it ranks first or ties for first in every column of Table~\ref{tab:main_results} against ten baselines and its VLA-Adapter backbone. It scores $76.5$ against $71.2$ on LIBERO, reaches $42.0$ on LIBERO-Long, $12.7$ points above the strongest baseline, scores $51.8$ against $45.4$ on LIBERO-Plus, where it wins $77$ more of the $1{,}200$ paired variants than its backbone, a significant margin (exact McNemar test), and raises real-robot success from $40.0\%$ to $60.0\%$. Neuro-symbolic structure~\citep{Neurosymbolic, Symbol-LLM} thus makes VLAs data-efficient, robust and interpretable on long-horizon manipulation.

\begin{table*}[t]
\centering
\caption{Related paradigms and our pilot results (colors as the dots): (a) LIBERO-Plus success rate (SR) per perturbation; (b) LIBERO-Long SR after the first primitive (chain) and of RL rollouts.}
\label{tab:related}
\newcommand{\pfam}[1]{\textcolor{#1}{\ensuremath{\bullet}}\,}
\newcommand{\form}[1]{$#1$}
\begin{minipage}[c]{\dimexpr\linewidth-116pt-7pt\relax}
\setlength{\tabcolsep}{1.3pt}
\fontsize{8pt}{10pt}\selectfont
\renewcommand{\arraystretch}{1.12}
\begin{tabular}{@{\hspace{1pt}}l<{\hspace{1.9pt}}l cccc@{\hspace{1pt}}}
\toprule
\textbf{Paradigm} & \textbf{Core formulation} & Ctrl. & Sym. & Comp. & Hier. \\
\midrule
\pfam{famimi}Imitation~\citep{robomimic, Diffusion-policy} & \form{\min_\theta \ell(\pi_\theta(o_t, x), a_t)} & \yes & \no & \no & \no \\
\pfam{famvla}Generalist VLA~\citep{OpenVLA, VLA-Adapter} & \form{A_t = f_\theta(I_t, S_t, x)} & \yes & \no & \no & \no \\
\pfam{famvla}VLA + online RL~\citep{VLA-RL, SimpleVLA-RL} & \form{\max_\theta \mathbb{E}[R(\tau)],\, R \in \{0, 1\}} & \yes & \no & \no & \no \\
\hspace{6pt}Hierarchical RL~\citep{The-option-critic-architecture, Data-Efficient-Hierarchical-Reinforcement-Learning} & \form{\max \mathbb{E}[e^{\mathrm{adv}/\beta} \log \pi(a{|}s,u)]} & \yes & \no & \no & \yes \\
\hspace{6pt}Neuro-symbolic~\citep{The-Neuro-Symbolic-Concept-Learner, NeSyPr} & \form{y = \mathrm{Exec}(\mathrm{Parse}(x))} & \no & \yes & \yes & \no \\
\midrule
\rowcolor{nslight}
\shadeL[1pt]\pfam{famns}\textbf{NS-VLA (ours)} & \form{\pi^{\mathrm{h}}_\phi(u_t{|}\mathcal{H}_t)\,\pi^{\mathrm{l}}_\theta(A_t{|}I_t,S_t,\tilde{x}_t)} & \yes & \yes & \yes & \yes\shadeR[1pt] \\
\bottomrule
\end{tabular}
\end{minipage}\hfill
\begin{minipage}[c]{116pt}
\includegraphics[width=116pt]{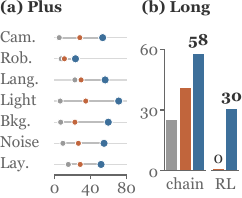}
\end{minipage}
\vspace{-6pt}
\end{table*}

\section{Related work}
\textbf{Neuro-symbolic learning.}
Neuro-symbolic methods pair neural perception with symbolic structure for compositional, verifiable reasoning, from concept learners and module networks~\citep{The-Neuro-Symbolic-Concept-Learner, Neural-Module-Networks-for-Reasoning-over-Text, Weakly-Supervised-Neural-Symbolic-Learning-for-Cognitive-Tasks} to probabilistic logic, theorem provers and program synthesis~\citep{DeepProbLog, Learning-knowledge-base-inference-with-neural-theorem-provers, DreamCoder}. LLMs extend them with programs, tools and solvers~\citep{PAL, Toolformer, ReAct, Logic-LM, Interactive-Evolution, Symbol-LLM}, and embodied agents compose skills, code and value maps from language~\citep{Do-As-I-Can-Not-As-I-Say, Code-as-Policies, Inner-Monologue, VoxPoser, pmlr, NeSyPr}. NS-VLA brings this structure into closed-loop control: a parsed plan and a gated stage pointer drive a chunked VLA policy, one primitive at a time.

\textbf{Vision-language-action models.}
Following generalist robot policies~\citep{RT-1, Octo}, VLAs fine-tune vision-language backbones~\citep{RT-2, OpenVLA, PI_0, pi_0.5} with latent, tokenized or world-model actions~\citep{UniVLA, FAST, WorldVLA}, efficient designs~\citep{RoboMamba, cogvla, A-Dual-Process-VLA, RoboFlamingo, Nora, OpenVLA-OFT} and compact backbones~\citep{VLA-Adapter, SmolVLA}, evaluated on LIBERO, LIBERO-Plus and CALVIN~\citep{LIBERO, LIBERO-Plus, CALVIN}. Online RL tunes them beyond demonstrations~\citep{VLA-RL, VLA-R1, SimpleVLA-RL, VLA-RFT, SRPO, EVOLVE-VLA, Interactive-Post-Training-for-Vision-Language-Action-Models} via policy-gradient and advantage-weighted updates~\citep{Proximal-Policy-Optimization-Algorithms, Advantage-Weighted-Regression, AWAC}, and hierarchical RL and VLAs add temporal abstraction~\citep{The-option-critic-architecture, Feudal-Reinforcement-Learning, Data-Efficient-Hierarchical-Reinforcement-Learning, LCB, VAMOS, VINE}. NS-VLA unites these lines in one hierarchical VLA.

\section{Preliminaries}
\label{sec:prelim}

\afterpage{%
\begin{figure*}[t]
\centering
\includegraphics[width=\linewidth]{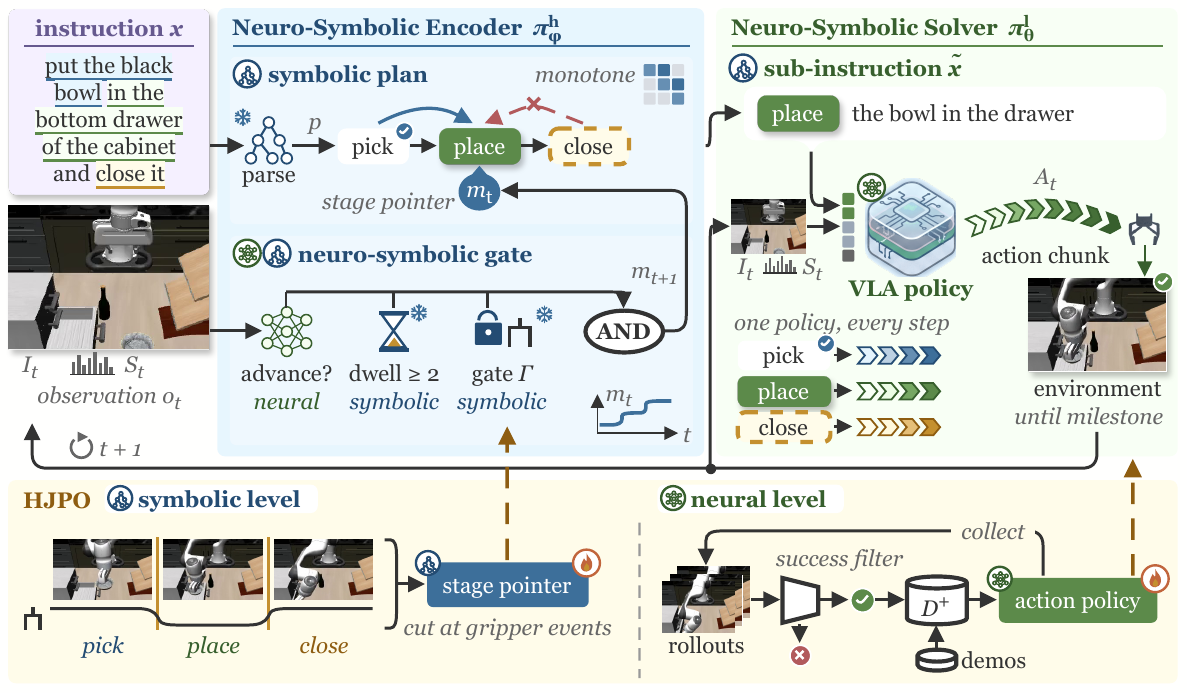}
\caption{Overview of NS-VLA: the Encoder parses a primitive plan and gates a monotone pointer, the Solver conditions the backbone on the active primitive, and HJPO trains both factors.}
\label{f4}
\end{figure*}}

\textbf{Problem formulation.}
Manipulation is a POMDP~\citep{Planning-and-acting-in-partially-observable-stochastic-domains}: at decision $t$ a VLA $\pi_\theta$ maps instruction $x$ and observation $o_t = (I_t, S_t)$ (images, proprioception) to an $H$-step chunk $A_t$. Given history $\mathcal{H}_t$, a neuro-symbolic VLA factorizes it into a symbolic high level for primitive $u_t$ and a neural low level:
\begin{equation}
\pi_{\Theta}\bigl(u_t, A_t \mid \mathcal{H}_t\bigr) = \pi^{\mathrm{h}}_\phi\bigl(u_t \mid \mathcal{H}_t\bigr) \cdot \pi^{\mathrm{l}}_\theta\bigl(A_t \mid u_t, \mathcal{H}_t\bigr), \qquad \Theta = (\phi, \theta).
\label{eq:hier-policy}
\end{equation}
Here $x$ is parsed into a primitive plan $p = (u^{(1)}, \ldots, u^{(M)})$ under pointer $m_t$, $u_t = u^{(m_t)}$; both factors learn from demonstrations, the low level also from rollouts, to maximize \mbox{$\mathbb{E}[R(\tau)]$}, \mbox{$R(\tau) = \mathbf{1}[\tau \models x]$}. This separates what to do from how to move and equals the VLA when $M = 1$.

\textbf{Pilot study.}
Table~\ref{tab:related} compares paradigms by continuous control (Ctrl.), symbolic structure (Sym.), primitive-level conditioning (Comp.) and hierarchical optimization (Hier.); only NS-VLA covers all four. The pilots confirm the challenges. (i) After the first primitive, imitation and generalist VLAs complete $25.0\%$ and $40.9\%$ of LIBERO-Long versus $57.9\%$ for NS-VLA. (ii) Monolithic families average below $35\%$ per Plus category, while NS-VLA leads all seven, five by over $25$ points. (iii) Flat RL on VLA-Adapter succeeds in none of $640$ rollouts, a silent reward, versus $195$ for HJPO.

\section{Method: NS-VLA}
\label{sec:method}

As shown in Figure~\ref{f4}, the Encoder realizes the symbolic high level $\pi^{\mathrm{h}}_\phi$, the Solver the neural low level $\pi^{\mathrm{l}}_\theta$, and HJPO trains both from demonstrations and the policy's own successes.

\subsection{Neuro-Symbolic Encoder}
\label{sec:ns_encoder}

\textbf{Compositional plan grammar.}
The Encoder compiles $x$ over the vocabulary $\mathcal{U} = \{$\texttt{pick}, \texttt{place}, \texttt{open}, \texttt{close}, \texttt{turn on}, \texttt{turn off}, \texttt{push}$\}$. A deterministic grammar $\mathcal{P}$ splits $x$ into clauses $c_1, \ldots, c_C$, resolves shared verbs and anaphora, keeps qualifiers and rewrites clauses by $\mathrm{rw}$:
\begin{equation}
p \;=\; \mathcal{P}(x) \;=\; \bigoplus\nolimits_{i=1}^{C} \mathrm{rw}(c_i), \qquad \mathrm{rw}\bigl(\texttt{put}(o, \rho, g)\bigr) \;=\; \bigl((\texttt{pick}, o),\, (\texttt{place}_{\rho}, (o, g))\bigr),
\label{eq:plan}
\end{equation}
where $\oplus$ concatenates, $o$ and $g$ are the object and its target, $\rho$ is the placement relation, and other clauses map to $(\mathrm{op}_c, \mathrm{arg}_c)$. Each $u$ is verbalized as a sub-instruction $\mathrm{text}(u)$, so $p = (u^{(1)}, \ldots, u^{(M)})$ is executable and human-readable, and an undecomposed instruction runs the one-slot plan $p = (x)$. Parsing runs once per episode, before the first action, and requires no learned parameters.

\textbf{Symbolic precondition gate.}
A primitive is complete only once its effect shows in the robot's own state. Let $\kappa_t \in \{0, 1\}$ be the last gripper command ($1$ = close), $w_t$ the finger separation in $S_t$ and $h_t = \kappa_t\, \mathbf{1}[w_t > w_{\min}]$ the holding indicator with threshold $w_{\min}$; $\mathcal{U}$ splits into grasp $\mathcal{U}_{\mathrm{g}} = \{\texttt{pick}\}$, grasp--release $\mathcal{U}_{\mathrm{r}}$ and gripper-free $\mathcal{U}_{\mathrm{f}}$ primitives, and $t_m$ is the first decision of slot $m$:
{\setlength{\abovedisplayskip}{8pt plus 2pt}\setlength{\abovedisplayshortskip}{8pt plus 2pt}%
\setlength{\belowdisplayskip}{0pt plus 2pt}\setlength{\belowdisplayshortskip}{0pt plus 2pt}%
\begin{equation}
\Gamma_t(u) \;=\; \mathbf{1}\bigl[u \in \mathcal{U}_{\mathrm{f}}\bigr] \;+\; \mathbf{1}\bigl[u \in \mathcal{U}_{\mathrm{g}}\bigr]\, h_t \;+\; \mathbf{1}\bigl[u \in \mathcal{U}_{\mathrm{r}}\bigr]\, (1 - \kappa_t) \smash[b]{\max_{t_{m_{t-1}} \leq t' \leq t}} h_{t'},
\label{eq:gate}
\end{equation}}%
so a \texttt{pick} is certified only while an object is held, a grasp--release one only after a grasp and a release in its slot, and a gripper-free one is left to the classifier. This closed-form predicate on proprioception and commands makes every transition auditable and independent of the images and the backbone.

\textbf{Gated monotone stage pointer.}
A light classifier $g_\phi$ judges stage completion from features $z_t$ fusing the pooled backbone context $\bar{h}_t$ of observation and instruction, embeddings $e(\cdot)$ of the current and next sub-instruction, the state $S_t$, the dwell time $d_t$ since the last advance and the progress $m_{t-1}/M$. The pointer then advances by at most one slot per decision, never skipping or revisiting a step:
\begin{equation}
m_t \;=\; m_{t-1} \;+\; \mathbf{1}\bigl[m_{t-1} < M\bigr]\; \mathbf{1}\bigl[d_t \geq d_{\min}\bigr]\; \Gamma_t\bigl(u^{(m_{t-1})}\bigr)\; \mathbf{1}\bigl[\sigma\bigl(g_\phi(z_t)\bigr) > \tfrac{1}{2}\bigr],
\label{eq:pointer}
\end{equation}
where $\sigma$ is the logistic function and $d_{\min}$ the minimum dwell. A neural judgment thus moves the pointer only when a symbolic rule agrees, and the high-level factor is the deterministic policy $\pi^{\mathrm{h}}_\phi(u \mid \mathcal{H}_t) = \mathbf{1}[u = u^{(m_t)}]$, whose trace can be read off and verified at every decision.

\begin{proposition}[Certified execution]
\label{prop:ns_consistency}
For every $\phi$, the pointer splits each episode into $\leq M$ ordered contiguous segments, each non-final one lasting $\geq d_{\min}$ decisions and left only once $\Gamma$ certifies it.
\end{proposition}
\vspace{-3mm}
\begin{proof}
See Appendix~\ref{proof1}; Section~\ref{sec:rq3} measures how well pointer advances are grounded.
\end{proof}

\newcommand{\slotop}[1]{\makebox[1.05cm][l]{\texttt{#1}}}
\newcommand{\Ops}[2]{\slotop{#1}\slotop{#2}}
\subsection{Neuro-Symbolic Solver}
\label{sec:solver}

\textbf{Primitive-conditioned chunk policy.}
The Solver grounds the active primitive in an unchanged chunked VLA backbone $f_\theta$. Its bridged input $\tilde{x}^{(m)} = x \oplus \texttt{now:} \oplus \mathrm{text}(u^{(m)})$ keeps $x$ and names the active step ($\tilde{x}^{(1)} = x$ for a one-slot plan), and $f_\theta$ is read as the mode of a Laplace chunk policy:
\begin{equation}
\pi^{\mathrm{l}}_\theta\bigl(A_t \mid u^{(m_t)}, \mathcal{H}_t\bigr) \;=\; (2b)^{-7H} \exp\!\Bigl(-\tfrac{1}{b}\, \bigl\| A_t - f_\theta\bigl(I_t,\, S_t,\, \tilde{x}^{(m_t)}\bigr) \bigr\|_1\Bigr),
\label{eq:action_chunk}
\end{equation}
where $A_t \in \mathbb{R}^{H \times 7}$ stacks $H$ end-effector deltas with gripper commands and $b > 0$ is a scale. The mode runs open-loop and is re-predicted right after an advance, so every executed chunk is conditioned on a certified primitive; L1 regression is exact maximum likelihood under this policy.

\textbf{Sub-instruction-augmented imitation.}
Demonstrations are segmented by gripper events at no annotation cost, so each frame $i$ carries its segment $m$ and chunk $A^\star_i$. A fair coin $\xi \sim \mathrm{Bern}(1/2)$ selects the input $x^{\xi}_i$, bridged $\tilde{x}^{(m)}$ if $\xi = 1$ and native $x$ otherwise, and the backbone minimizes
\begin{equation}
\mathcal{L}_{\mathrm{SFT}}(\theta) \;=\; \mathbb{E}_{(x,\, o_i,\, A^\star_i,\, m) \sim \mathcal{D},\;\, \xi \sim \mathrm{Bern}(1/2)}\Bigl[\bigl\| f_\theta\bigl(I_i,\, S_i,\, x^{\xi}_i\bigr) - A^\star_i \bigr\|_1\Bigr],
\label{eq:sft}
\end{equation}
which keeps the backbone fluent on native instructions while every segment adds a primitive-specific signal. Because sub-instructions recur across tasks, a primitive learned in one chain is rehearsed whenever it reappears in another, which lets limited demonstrations cover long horizons.

\textbf{Cross-task primitive sharing.}
This sharing is exact. Let $\mathcal{L}_{\mathrm{nat}}$ be the imitation loss on native inputs, $\lambda_u$ the fraction of demonstration frames whose segment executes $u$ ($u = x$ on one-slot plans), and $\mathcal{L}_u$ the mean bridged loss over those frames pooled over tasks; as $\xi$ is independent of the data,
\begin{equation}
\mathcal{L}_{\mathrm{SFT}}(\theta) \;=\; \tfrac{1}{2}\, \mathcal{L}_{\mathrm{nat}}(\theta) \;+\; \tfrac{1}{2} \sum_{u} \lambda_u\, \mathcal{L}_u(\theta), \qquad \sum_{u} \lambda_u = 1,
\label{eq:sft-share}
\end{equation}
so half of the supervision is organized by primitive rather than by task, and each $\mathcal{L}_u$ aggregates every demonstration of $u$ in the suite, whatever task it came from. On a one-slot plan the two halves coincide frame by frame, and the objective is exactly the backbone's imitation loss.

\begin{proposition}[Graceful equivalence]
\label{prop:graceful_equiv}
If $M = 1$, then for every $\theta$ NS-VLA equals its backbone $\pi_\theta(\cdot \mid I_t, S_t, x)$ at every history, with the same trajectory law and training objectives.
\end{proposition}
\vspace{-3mm}
\begin{proof}
See Appendix~\ref{proof2}; Section~\ref{sec:rq4} studies the Solver on multi-primitive LIBERO-Long plans.
\end{proof}

\subsection{Hierarchical Joint Policy Optimization}
\label{sec:online_rl}

\textbf{Factorized hierarchical likelihood.}
HJPO trains both factors with signals matched to their types, discrete stage transitions for the Encoder and continuous action chunks for the Solver. Viewing each stay/advance decision $y_t$ as a Bernoulli variable with $p_\phi(y_t = 1 \mid z_t) = \sigma(g_\phi(z_t))$, the log-likelihood of a rollout $\tau$ under the hierarchical policy separates into a discrete and a continuous part:
\begin{equation}
\log \pi_\Theta(\tau) \;=\; \sum_{t=1}^{T} \Bigl[\log p_\phi\bigl(y_t \mid z_t\bigr) \;+\; \log \pi^{\mathrm{l}}_\theta\bigl(\hat{A}_t \mid u^{(m_t)}, \mathcal{H}_t\bigr)\Bigr] \;+\; \log \mu_{\mathrm{env}}(\tau),
\label{eq:hjpo-lik}
\end{equation}
where $\hat{A}_t$ is the executed chunk and $\mu_{\mathrm{env}}$ collects the parameter-free initial-state and transition terms. With $z_t$ as fixed inputs, the gradient in $\phi$ involves only stage transitions and the gradient in $\theta$ only action chunks, so each factor fits the signal suited to it, without a learned critic or value function.

\textbf{High-level factor.}
The classifier learns from segments: each frame of segment $j$ yields a stay example under pointer $j$ and, for $j > 1$, an advance example under pointer $j - 1$, fit by class-balanced cross-entropy. With the gate it fixes $\pi^{\mathrm{h}}_\phi$ and each rollout's sub-instructions for online refinement.

\textbf{Low-level factor.}
The action policy is first fit by sub-instruction-augmented imitation and then improved online from its own experience. Each iteration rolls out the current hierarchical policy with exploration noise, stores every executed chunk of each successful rollout with its bridged input in a replay set $\mathcal{D}^{+}$, and mixes it with the expert demonstrations at a fixed mixing ratio $\eta$:
\begin{equation}
\mathcal{L}_{\mathrm{HJPO}}(\theta) \;=\; \eta\, \mathcal{L}_{\mathrm{SFT}}(\theta) \;+\; (1 - \eta)\; \mathbb{E}_{(o,\, \tilde{x},\, \hat{A}) \sim \mathcal{D}^{+}} \bigl\| f_\theta(I, S, \tilde{x}) - \hat{A} \bigr\|_1.
\label{eq:bar}
\end{equation}
This is advantage-weighted regression (AWR)~\citep{Advantage-Weighted-Regression} at zero temperature with task-level credit: the Laplace log-likelihood of the Solver turns its weighted maximum likelihood exactly into L1 regression, and under the binary reward its weights collapse to a hard, parameter-free success filter,
\begin{equation}
w_\beta(\tau) \;=\; \exp\!\Bigl(\frac{R(\tau) - R_{\max}}{\beta}\Bigr) \;\xrightarrow[\;\beta \to 0\;]{}\; \mathbf{1}\bigl[R(\tau) = R_{\max}\bigr], \qquad R_{\max} = 1,
\label{eq:awr-limit}
\end{equation}
with temperature $\beta > 0$, so each update regresses the policy onto its own successes, anchored by the demonstrations, while failed rollouts drop out of the update without any reward shaping. The reported policy $\hat{\theta}$ maximizes the validation success rate $\mathrm{SR}_{\mathrm{val}}$ over the validated iterates, the supervised start $\theta_0$ itself included, so selection never falls below the supervised policy on validation.

\begin{proposition}[Filtered AWR]
\label{prop:rl_convergence}
With binary reward and nonempty $\mathcal{D}^{+}$, as $\beta \to 0$ anchored AWR converges to $\mathcal{L}_{\mathrm{HJPO}}$ and to the success-conditioned law at rate $e^{-1/\beta}$, and $\mathrm{SR}_{\mathrm{val}}(\hat{\theta}) \geq \mathrm{SR}_{\mathrm{val}}(\theta_0)$.
\end{proposition}
\vspace{-3mm}
\begin{proof}
See Appendix~\ref{proof3}; Section~\ref{sec:rq5} reports the online stage across all 19 online runs.
\end{proof}

\begin{table*}[!t]
\centering
\caption{SR (\%) on LIBERO and LIBERO-Plus (Plus Avg pools 1,200 variants); subscripts: standard error (no success: 95\% bound $u$ as $u/2\pm u/2$); \textbf{best}, \underline{second}; $\Delta$: gain over best baseline.}
\label{tab:main_results}
\setlength{\tabcolsep}{2.26pt}
\fontsize{8pt}{10pt}\selectfont
\begin{tabular}{@{\hspace{0.5pt}}l@{\hspace{2pt}}ccccc|cccccccc@{\hspace{0.5pt}}}
\toprule
\multirow{2}{*}{\textbf{Method}}
& \multicolumn{5}{c|}{\textbf{LIBERO}}
& \multicolumn{8}{c}{\textbf{LIBERO-Plus}} \\
& \colhead{Spat.} & \colhead{Obj.} & \colhead{Goal} & \colhead{Long} & \colhead{Avg}
& \colhead{Cam.} & \colhead{Rob.} & \colhead{Lang.} & \colhead{Light} & \colhead{Bkg.} & \colhead{Noise} & \colhead{Lay.} & \colhead{Avg} \\
\midrule
BC-RNN & 7.0\se{1.8} & 21.5\se{2.9} & 13.5\se{2.4} & 8.5\se{2.0} & 12.6\se{1.2} & 0.8\se{0.8} & 0.5\se{0.5} & 3.8\se{1.4} & 0.7\se{0.7} & 1.1\se{1.1} & 0.5\se{0.5} & 2.2\se{1.1} & 1.2\se{0.3} \\
BC-ViLT & 26.0\se{3.1} & 64.5\se{3.4} & 35.0\se{3.4} & 3.5\se{1.3} & 32.3\se{1.5} & 0.8\se{0.8} & 1.1\se{0.8} & 10.8\se{2.3} & 6.6\se{2.1} & 1.6\se{1.1} & 0.8\se{0.8} & 3.8\se{1.4} & 3.3\se{0.5} \\
QueST & 46.0\se{3.5} & 65.5\se{3.4} & 47.0\se{3.5} & 16.5\se{2.6} & 43.8\se{1.6} & 11.0\se{2.3} & 10.2\se{2.2} & 31.9\se{3.4} & 11.0\se{2.7} & 14.7\se{3.1} & 14.7\se{2.6} & 24.2\se{3.2} & 17.1\se{1.1} \\
BC-T & 48.5\se{3.5} & 63.0\se{3.4} & 47.5\se{3.5} & 16.5\se{2.6} & 43.9\se{1.6} & 0.8\se{0.8} & 4.3\se{1.5} & 11.4\se{2.3} & 2.9\se{1.4} & 3.1\se{1.5} & 0.5\se{0.5} & 7.7\se{2.0} & 4.3\se{0.6} \\
VQ-BeT & 44.0\se{3.5} & 71.5\se{3.2} & 51.5\se{3.5} & 15.5\se{2.6} & 45.6\se{1.6} & 5.8\se{1.7} & 7.0\se{1.9} & 20.5\se{3.0} & 0.7\se{0.7} & 8.5\se{2.5} & 8.9\se{2.1} & 15.9\se{2.7} & 10.0\se{0.8} \\
Smol. & 57.5\se{3.5} & 59.0\se{3.5} & 52.5\se{3.5} & 27.5\se{3.2} & 49.1\se{1.7} & 0.8\se{0.8} & 0.8\se{0.8} & 3.2\se{1.3} & 5.1\se{1.9} & 5.4\se{2.0} & 0.5\se{0.5} & 2.2\se{1.1} & 2.1\se{0.4} \\
Octo & 61.5\se{3.4} & 66.5\se{3.3} & 60.5\se{3.5} & 11.5\se{2.3} & 50.0\se{1.6} & 33.0\se{3.4} & 12.9\se{2.5} & 36.2\se{3.5} & 39.7\se{4.2} & 6.2\se{2.1} & 35.6\se{3.5} & 34.6\se{3.5} & 28.9\se{1.2} \\
DP & 59.0\se{3.5} & 77.5\se{3.0} & 57.5\se{3.5} & 29.0\se{3.2} & 55.8\se{1.6} & 2.1\se{1.0} & 8.6\se{2.1} & 21.1\se{3.0} & 7.4\se{2.2} & 8.5\se{2.5} & 12.0\se{2.4} & 21.4\se{3.0} & 11.8\se{0.9} \\
BAKU & 38.5\se{3.4} & \underline{85.5}\se{2.5} & \textbf{81.0}\se{2.8} & 24.5\se{3.0} & 57.4\se{1.5} & 19.9\se{2.9} & 18.3\se{2.8} & 39.5\se{3.6} & 14.7\se{3.0} & 7.8\se{2.4} & 23.6\se{3.1} & 31.3\se{3.4} & 23.1\se{1.2} \\
ACT & 79.5\se{2.9} & 78.5\se{2.9} & 62.0\se{3.4} & 20.0\se{2.8} & 60.0\se{1.5} & 1.0\se{0.7} & 7.0\se{1.9} & 40.5\se{3.6} & 4.4\se{1.8} & 8.5\se{2.5} & 12.6\se{2.4} & 14.3\se{2.6} & 13.1\se{0.9} \\
VLA-Ad. & \underline{90.0}\se{2.1} & \textbf{87.0}\se{2.4} & \underline{78.5}\se{2.9} & \underline{29.3}\se{14.7} & \underline{71.2}\se{3.8} & \underline{50.8}\se{3.6} & \underline{19.4}\se{2.9} & \underline{48.6}\se{3.7} & \underline{58.8}\se{4.2} & \underline{55.8}\se{4.4} & \underline{42.9}\se{3.6} & \underline{48.4}\se{3.7} & \underline{45.4}\se{1.2} \\
\midrule
\rowcolor{nslight}
\shadeL[0.5pt]\textbf{NS-VLA} & \textbf{96.0}\se{1.4} & \textbf{87.0}\se{2.4} & \textbf{81.0}\se{2.8} & \textbf{42.0}\se{3.5} & \textbf{76.5}\se{1.3} & \textbf{53.4}\se{3.6} & \textbf{23.1}\se{3.1} & \textbf{56.2}\se{3.6} & \textbf{71.3}\se{3.9} & \textbf{59.7}\se{4.3} & \textbf{55.0}\se{3.6} & \textbf{51.6}\se{3.7} & \textbf{51.8}\se{1.4}\shadeR[0.5pt] \\
\rowcolor{nslight}
\shadeL[0.5pt]$\Delta$ & \sdelta{+6.0} & \sdelta{+0.0} & \sdelta{+0.0} & \sdelta{+12.7} & \sdelta{+5.3} & \sdelta{+2.6} & \sdelta{+3.7} & \sdelta{+7.6} & \sdelta{+12.5} & \sdelta{+3.9} & \sdelta{+12.1} & \sdelta{+3.2} & \sdelta{+6.4}\shadeR[0.5pt] \\
\bottomrule
\end{tabular}
\end{table*}

\section{Experiments}
\label{sec:experiments}

We study seven research questions.
\textbf{RQ1:} Does NS-VLA beat policies trained on the same data?
\textbf{RQ2:} Does each component contribute?
\textbf{RQ3--RQ5:} How do the Encoder, Solver and HJPO each help?
\textbf{RQ6:} How does NS-VLA run a long instruction?
\textbf{RQ7:} Do gains transfer to real robots?

\subsection{Setup}
\label{sec:setup}

\textbf{Datasets.} LIBERO~\citep{LIBERO} has four ten-task suites: Spatial, Object, Goal and long-horizon Long. Every method trains on 10\% of the official demonstrations, the first five per task. LIBERO-Plus~\citep{LIBERO-Plus} is held out for evaluation: 10{,}030 variants of the same tasks along seven perturbation dimensions with five difficulty levels: camera (Cam.), robot initial state (Rob.), language (Lang.), lighting (Light), background (Bkg.), sensor noise (Noise) and object layout (Lay.). Details are in Appendix~\ref{sec:dataset_details}.

\textbf{Baselines.} We compare against ten language-conditioned policies with at most 2B parameters, each trained on the same five demonstrations per task with its official recipe: BC-RNN~\citep{robomimic}, BC-Transformer (BC-T) and BC-ViLT~\citep{LIBERO}, ACT~\citep{Learning-Fine-Grained-Bimanual-Manipulation-with-Low-Cost-Hardware}, Diffusion Policy (DP)~\citep{Diffusion-policy}, VQ-BeT~\citep{VQ-BeT}, QueST~\citep{QueST}, BAKU~\citep{BAKU}, Octo~\citep{Octo} and SmolVLA (Smol.)~\citep{SmolVLA}, together with VLA-Adapter (VLA-Ad.)~\citep{VLA-Adapter}, the backbone that NS-VLA builds on. Further details are in Appendix~\ref{sec:baseline_details}.

\textbf{Metrics.} Success rate (SR) is the share of successful episodes. LIBERO uses the first 20 official initial states per task and LIBERO-Plus 300 pre-registered variants per suite, shared by all methods; RQ2--RQ5 mainly analyze the 500 LIBERO-Long test episodes, which chain primitives. The Plan-Stage Completion Rate (PSCR) is the share of plan stages whose simulator milestone is reached.

\textbf{Implementation.} NS-VLA wraps VLA-Adapter (Qwen2.5-0.5B with DINOv2 and SigLIP encoders) initialized from the generic vision-language model. Supervised training runs 20k steps on Long and 10k elsewhere, and the online stage runs 8 iterations of 8 rollouts per task. The reported model is selected on validation states, the supervised start included. Hyperparameters are in Appendix~\ref{sec:implementation_details}.

\subsection{Main results (RQ1)}
\label{sec:rq1}

As shown in Table~\ref{tab:main_results}, we compare NS-VLA with ten baselines and its VLA-Adapter backbone on LIBERO and the held-out LIBERO-Plus, and it ranks first or ties for first in all thirteen columns.

\textbf{The lead is largest where tasks chain primitives.} On LIBERO, NS-VLA averages $76.5$ against $71.2$ for VLA-Adapter and $60.0$ for ACT, and on Long it reaches $42.0$, $12.7$ points above the strongest baseline, while no baseline exceeds $30$; this gain is more than twice the gain on any other LIBERO suite. On Spatial it reaches $96.0$, six points above VLA-Adapter, the best baseline, and on Object and Goal it matches the strongest baselines, VLA-Adapter and BAKU, at $87.0$ and $81.0$.

\begin{figure*}[t]
\centering
\includegraphics[width=\textwidth]{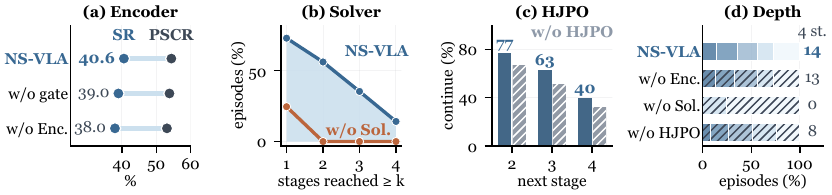}
\caption{Ablations: (a) full model, without gate and without Encoder; (b) episodes reaching at least $k$ stages; (c) share of episodes continuing to the next stage; (d) plan stages reached per variant.}
\label{fig:rq2_ablation}
\vspace{-6pt}
\end{figure*}

\textbf{The lead holds under perturbation.} On the held-out LIBERO-Plus, NS-VLA averages $51.8$ against $45.4$ for VLA-Adapter and $28.9$ for Octo and leads all seven dimensions, with the largest margins under lighting ($+12.5$), sensor noise ($+12.1$) and language ($+7.6$). Over the $1{,}200$ paired variants it wins $77$ more variants than VLA-Adapter, a significant margin (exact McNemar test, $p = 1.5\times10^{-5}$). Its margin over VLA-Adapter thus grows from $5.3$ points on LIBERO to $6.4$ points under perturbation.

\subsection{Ablation study (RQ2)}
\label{sec:rq2}

\begin{wraptable}{r}{0.4\linewidth}
\vspace{-13pt}
\centering
\caption{Ablations of each component.}
\label{tab:ablation}
\fontsize{8pt}{10pt}\selectfont
\setlength{\tabcolsep}{0.8pt}
\begin{tabular}{@{}l>{\hspace{8pt}}rl>{\hspace{8pt}}rl@{}}
\toprule
\textbf{Variant} & \multicolumn{2}{c}{\textbf{SR}} & \multicolumn{2}{c}{\textbf{PSCR}} \\
\midrule
\rowcolor{nslight}
\shadeL\textbf{NS-VLA} & \textbf{40.6} & \phantom{\sdrop{40.6}} & \textbf{54.4} & \phantom{\sdrop{47.1}}\shadeR \\
w/o Encoder & 38.0 & \sdrop{2.6} & 53.1 & \sdrop{1.3} \\
w/o Solver  & 0.0 & \sdrop{40.6} & 7.3 & \sdrop{47.1} \\
w/o HJPO    & 28.2 & \sdrop{12.4} & 48.9 & \sdrop{5.5} \\
\bottomrule
\end{tabular}
\vspace{-15.5pt}
\end{wraptable}
We remove one component at a time: Table~\ref{tab:ablation} reports SR and PSCR, with drops as subscripts, and Figure~\ref{fig:rq2_ablation} shows at which plan stage each component acts.

\textbf{Every component contributes.} Removing any component lowers both SR and PSCR. The Solver lifts SR from $0.0$ to $40.6$, HJPO adds $12.4$ points over supervised training, and the Encoder another $2.6$ points at inference.

\textbf{Each component plays a distinct role.} The gate and the Encoder each add success, lifting SR from $39.0$ and $38.0$ to $40.6$ (a). With the Solver, $56.2\%$ of episodes complete two stages, which the backbone never does (b). HJPO raises continuation at every later stage to $77$, $63$ and $40\%$ (c), so the full model reaches four stages in $14.2\%$ of episodes, more often than any ablated variant (d).

\subsection{Analysis of the Neuro-Symbolic Encoder (RQ3)}
\label{sec:rq3}

\begin{figure}[!h]
\centering
\includegraphics[width=\textwidth]{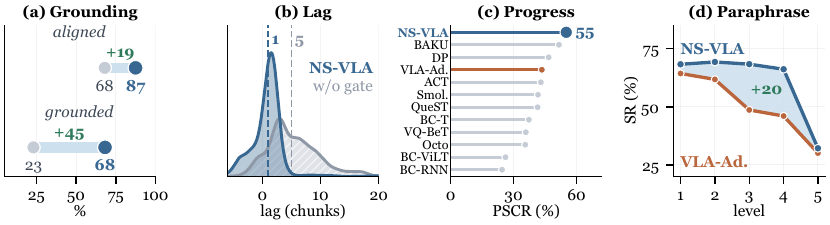}
\caption{Encoder: (a) the gate grounds pointer advances in milestones; (b) lag from an advance to its milestone; (c) PSCR of all twelve methods; (d) SR on LIBERO-Plus paraphrases by level.}
\label{fig:rq3_encoder}
\vspace{-6pt}
\end{figure}

As shown in Figure~\ref{fig:rq3_encoder}, we test whether the plan and the monotone pointer track execution, measuring how well pointer advances match simulator milestones and how the plan transfers to paraphrases.

\textbf{The precondition gate grounds the pointer.} Of NS-VLA's advances, $68.2\%$ see their milestone at most three chunks later, against $23.2\%$ without the gate, and the final pointer lies within one stage of the achieved progress in $87.4\%$ of episodes (a); median lag falls from five chunks to one (b).

\textbf{Structured execution reaches further and absorbs paraphrases.} NS-VLA attains the highest PSCR of all twelve methods, $55.1$ against $51.6$ for BAKU, so it completes the most plan stages (c). On reworded LIBERO-Plus instructions it stays above its VLA-Adapter backbone at every difficulty level, by up to $20.2$ points, and keeps its SR at $66$--$69\%$ through level 4 of the five levels (d).

\begin{figure*}[t]
\centering
\includegraphics[width=\textwidth]{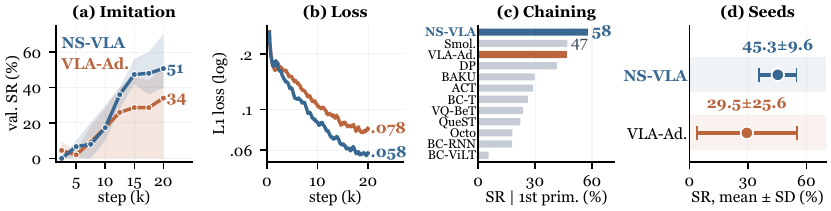}
\caption{Solver: (a) validation SR (band: range) and (b) training loss over three seeds; (c) task SR once the first primitive is done, for all methods; (d) test SR over three training seeds.}
\label{fig:rq4_solver}
\vspace{-6pt}
\end{figure*}

\subsection{Analysis of the Neuro-Symbolic Solver (RQ4)}
\label{sec:rq4}

As shown in Figure~\ref{fig:rq4_solver}, we test whether sub-instructions make five demonstrations easier to imitate, following imitation and loss over three seeds and each method's success after the first primitive.

\textbf{The Solver learns faster and further.} Over three training seeds, NS-VLA's checkpoints lead from 12.5k steps on and finish at $51\%$ validation SR against $34\%$ (a), with lower training loss after warmup, $0.058$ against $0.078$ (b), as the Solver pools the segments of each primitive across all ten tasks.

\textbf{It chains primitives stably.} Once the first primitive is done, NS-VLA completes $57.9\%$ of episodes, ahead of SmolVLA ($47.0\%$) and every other compared method (c), and over all three training seeds it averages $45.3\pm9.6$ against $29.5\pm25.6$ for its VLA-Adapter backbone (d), so its standard deviation across these seeds is well under half that of the backbone on the same test episodes.

\subsection{Analysis of Hierarchical Joint Policy Optimization (RQ5)}
\label{sec:rq5}

\begin{figure}[!h]
\centering
\includegraphics[width=\textwidth]{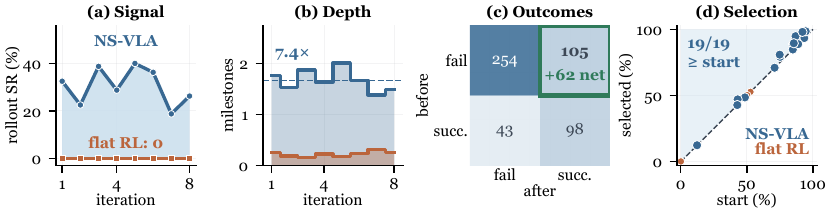}
\caption{Online stage of HJPO: (a) rollout SR and (b) milestones per rollout in every iteration; (c) test outcomes before and after the stage; (d) selected against starting validation SR of all runs.}
\label{fig:rq5_rl}
\vspace{-6pt}
\end{figure}

As shown in Figure~\ref{fig:rq5_rl}, we test whether the hierarchy turns online interaction into a learning signal, following every online iteration and the validation-based model selection over all online runs.

\begin{figure*}[t]
\centering
\includegraphics[width=\textwidth]{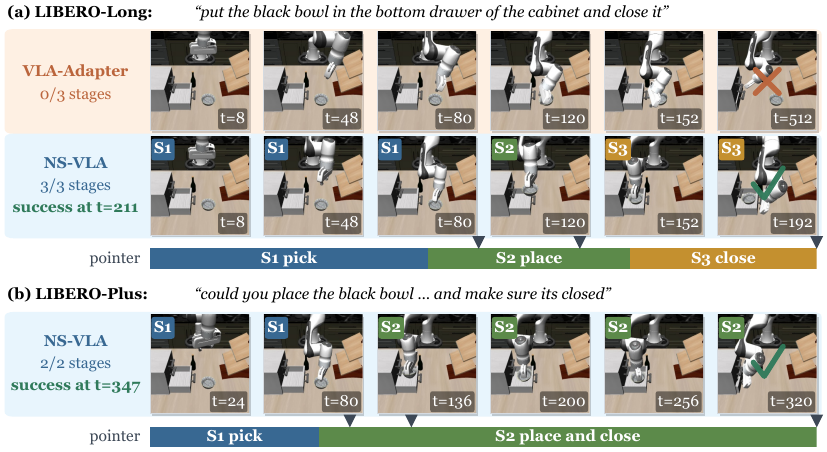}
\caption{Case study from the same initial state on (a) the original instruction and (b) a paraphrase: badges give the active plan stage S$k$, the bar shows the pointer over time and $\blacktriangledown$ marks milestones.}
\label{fig:rq6_case}
\vspace{-6pt}
\end{figure*}

\textbf{Structure makes rewards reachable.} NS-VLA succeeds in $18.8$--$40.0\%$ of its rollouts in every iteration, so each filtered update has successes to learn from, whereas flat RL on VLA-Adapter collects none (a), and NS-VLA's rollouts reach $1.67$ milestones on average, $7.4\times$ the flat baseline (b).

\textbf{The online stage converts failures safely.} On the 500 test episodes it turns 105 failures into successes, a net gain of 62 that lifts SR from $28.2$ to $40.6$ (c), and across all 19 online runs the selected model never falls below its supervised start on validation and gains up to 8 points over it (d).

\subsection{Case study (RQ6)}
\label{sec:rq6}

As shown in Figure~\ref{fig:rq6_case}, we replay the long-horizon task ``put the black bowl in the bottom drawer of the cabinet and close it'' and a LIBERO-Plus paraphrase of it from the same initial state.

\textbf{NS-VLA executes the plan stage by stage.} The plan has three stages, S1 ``pick up the black bowl'', S2 ``place the black bowl in the bottom drawer of the cabinet'' and S3 ``close the bottom drawer of the cabinet''. NS-VLA picks up the bowl, places it in the drawer and closes it, advancing within two chunks of each milestone, and succeeds at step 211, while VLA-Adapter reaches no stage (a).

\textbf{The plan carries over to a paraphrase.} The reworded instruction ``could you place the black bowl \ldots\ and make sure its closed'' is parsed into a pick stage and a place stage. The pointer leaves the pick stage as soon as the gripper holds the bowl, and NS-VLA succeeds at step 347 (b), so the plan absorbs the rewording without any retraining of the grammar parser or the action policy.

\subsection{Performance on real-world tasks (RQ7)}
\label{sec:rq7}

\begin{wrapfigure}{r}{0.5\textwidth}
\vspace{-12pt}
\centering
\includegraphics[width=\linewidth]{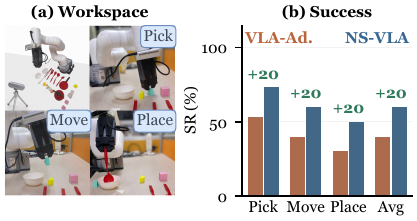}
\caption{Real robot: (a) setup; (b) success rate.}
\label{fig:rq7_real}
\vspace{-16pt}
\end{wrapfigure}
As shown in Figure~\ref{fig:rq7_real}, we deploy NS-VLA on a myCobot 320pi arm against VLA-Adapter to test whether the gains we observe in simulation also transfer to real hardware.

\textbf{The tasks span single and compositional skills.} With a wrist and a third-person RealSense D435i camera, both policies run the same three Pick, three Move and four Place tasks, each repeated ten times under randomized object placements, for 100 trials per policy in total (a).

\textbf{The gains transfer to the real robot.} NS-VLA reaches $73.3$ on Pick, $60.0$ on Move and $50.0$ on Place, ahead of VLA-Adapter by $20.0$ points in every category, and averages $60.0$ over all 100 trials against $40.0$, a relative gain of $50\%$, with the compositional Place tasks, which chain pick and place, keeping the full margin (b).

\section{Conclusion}
We presented NS-VLA, a neuro-symbolic vision-language-action framework that exposes the compositional structure of manipulation instructions to a chunked action policy. The Encoder compiles each instruction into a primitive plan tracked by a precondition-gated monotone stage pointer, the Solver conditions an unchanged backbone on the active sub-instruction, and HJPO fits the pointer on demonstration segments and refines the action policy by success-filtered advantage-weighted regression. By construction, every stage transition is certified, one-slot plans reduce exactly to the backbone, and online refinement never falls below its supervised start on validation. Empirically, NS-VLA ranks first or ties for first in every column of LIBERO and LIBERO-Plus, with the largest margin on long-horizon tasks, and raises real-robot success from $40.0\%$ to $60.0\%$. Explicit symbolic structure thus offers an effective route to data-efficient, robust and interpretable VLA policies.

\begin{ack}
This work was supported by the National Natural Science Foundation of China (Grant No.~62472043). We thank the anonymous reviewers and area chairs for their constructive feedback, which helped improve the clarity and rigor of this paper. The authors declare no competing interests.
\end{ack}

\bibliographystyle{plainnat}
\bibliography{custom}

\clearpage

\appendix
\raggedbottom

\section{Plan grammar and demonstration segmentation}
\label{sec:plan_grammar}

The Encoder maps an instruction to a primitive plan with a deterministic grammar, renders each primitive as a sub-instruction, and segments demonstrations into stages by gripper events. None of the three requires annotation; Spatial, Object and Goal run the one-slot plan (Figure~\ref{fig:app_prompts}).

\begin{figure}[!htb]
\centering
\begin{minipage}[t]{0.49\textwidth}
\begin{nsbox}{Encoder: instruction $\rightarrow$ plan}{nstop}
instruction turn on the stove and
            put the moka pot on it
clauses     [turn on the stove]
            [put the moka pot on it]
anaphora    it -> the stove
plan        (turn on, stove)
            (pick, moka pot)
            (place_on, (moka pot, stove))
\end{nsbox}
\begin{nsbox}{Rendering of a primitive}{nsbot}
(pick, o)          pick up the o
(place_rho, (o,g)) place the o rho the g
other primitives   normalized clause
\end{nsbox}
\end{minipage}\hfill
\begin{minipage}[t]{0.49\textwidth}
\begin{nsbox}{Solver: bridged input per stage}{nstop}
[S1] put the black bowl in the bottom
     drawer of the cabinet and close it.
     now: pick up the black bowl
[S2] ... now: place the black bowl in
     the bottom drawer of the cabinet
[S3] ... now: close the bottom drawer
     of the cabinet
\end{nsbox}
\begin{nsbox}{One-slot plan (M = 1)}{nsbot}
S/O/G: pick up the alphabet soup and
place it in the basket -> x itself
\end{nsbox}
\end{minipage}
\caption{Inputs of the Encoder and the Solver: a plan compiled from an instruction, the rendering of primitives as sub-instructions, and the bridged input the backbone reads at every stage.}
\label{fig:app_prompts}
\end{figure}

\subsection{Grammar and primitive vocabulary}

\textbf{Clause splitting.} The parser lower-cases the instruction, splits it at coordinating conjunctions and normalizes each clause to a verb frame. Three rewrites make the clauses self-contained. A shared verb is distributed over coordinated objects, and a plural object is expanded into its instances (``put both moka pots on the stove'' yields one clause per pot). A pronoun or elided target takes the most recent compatible referent (``\ldots in the microwave and close it'' $\to$ ``close the microwave''). Spatial qualifiers of object and target are kept verbatim (``the black bowl between the plate and the ramekin'').

\textbf{Vocabulary.} A clause $\texttt{put}(o, \rho, g)$ becomes the pair $(\texttt{pick}, o)$, $(\texttt{place}_\rho, (o, g))$, with $\rho$ one of \emph{on}, \emph{in}, \emph{on top of} or a spatial relation such as \emph{to the right of}. Every other clause yields one primitive, \texttt{open}, \texttt{close}, \texttt{turn on}, \texttt{turn off} or \texttt{push}, whose argument is the articulated object and, for \texttt{push}, the goal. The vocabulary is read off the training instructions, and objects remain free-text arguments. It is partitioned by the gripper behavior that the precondition gate of the Encoder checks:
\begin{center}
\fontsize{8pt}{10pt}\selectfont
\setlength{\tabcolsep}{5pt}
\begin{tabular}{@{}lll@{}}
\toprule
\textbf{Class} & \textbf{Primitives} & \textbf{Gripper signature} \\
\midrule
$\mathcal{U}_{\mathrm{g}}$ (grasp) & \texttt{pick} & ends with an object held \\
$\mathcal{U}_{\mathrm{r}}$ (grasp--release) & \texttt{place}, \texttt{turn on}, \texttt{turn off} & a grasp followed by a release \\
$\mathcal{U}_{\mathrm{f}}$ (gripper-free) & \texttt{open}, \texttt{close}, \texttt{push} & no gripper event \\
\bottomrule
\end{tabular}
\end{center}

\textbf{Rendering and bridged input.} A primitive is rendered as ``pick up the $o$'', ``place the $o$ $\rho$ the $g$'' or its normalized clause, and the Solver reads ``\texttt{$x$. now: $\mathrm{text}(u^{(m)})$}''; a one-slot plan reads $x$ itself. Table~\ref{tab:long_plans} lists every LIBERO-Long plan, fixed across training, rollouts and evaluation.

\begin{table}[!htb]
\centering
\caption{Plans of the ten LIBERO-Long instructions; ``$\mid$'' separates consecutive sub-instructions.}
\label{tab:long_plans}
\vspace{2pt}
\fontsize{8pt}{10pt}\selectfont
\setlength{\tabcolsep}{3pt}
\begin{tabular}{@{}c>{\raggedright\arraybackslash}p{0.33\linewidth}>{\raggedright\arraybackslash}p{0.57\linewidth}@{}}
\toprule
\textbf{Task} & \textbf{Instruction} & \textbf{Plan (sub-instructions)} \\
\midrule
0 & put both the alphabet soup and the tomato sauce in the basket & pick up the alphabet soup $\mid$ place the alphabet soup in the basket $\mid$ pick up the tomato sauce $\mid$ place the tomato sauce in the basket \\
1 & put both the cream cheese box and the butter in the basket & pick up the cream cheese box $\mid$ place the cream cheese box in the basket $\mid$ pick up the butter $\mid$ place the butter in the basket \\
2 & turn on the stove and put the moka pot on it & turn on the stove $\mid$ pick up the moka pot $\mid$ place the moka pot on the stove \\
3 & put the black bowl in the bottom drawer of the cabinet and close it & pick up the black bowl $\mid$ place the black bowl in the bottom drawer of the cabinet $\mid$ close the bottom drawer of the cabinet \\
4 & put the white mug on the left plate and put the yellow and white mug on the right plate & pick up the white mug $\mid$ place the white mug on the left plate $\mid$ pick up the yellow and white mug $\mid$ place the yellow and white mug on the right plate \\
5 & pick up the book and place it in the back compartment of the caddy & pick up the book $\mid$ place the book in the back compartment of the caddy \\
6 & put the white mug on the plate and put the chocolate pudding to the right of the plate & pick up the white mug $\mid$ place the white mug on the plate $\mid$ pick up the chocolate pudding $\mid$ place the chocolate pudding to the right of the plate \\
7 & put both the alphabet soup and the cream cheese box in the basket & pick up the alphabet soup $\mid$ place the alphabet soup in the basket $\mid$ pick up the cream cheese box $\mid$ place the cream cheese box in the basket \\
8 & put both moka pots on the stove & pick up the moka pot $\mid$ place the moka pot on the stove $\mid$ pick up the other moka pot $\mid$ place the other moka pot on the stove \\
9 & put the yellow and white mug in the microwave and close it & pick up the yellow and white mug $\mid$ place the yellow and white mug in the microwave $\mid$ close the microwave \\
\bottomrule
\end{tabular}
\end{table}

\subsection{Demonstration segmentation}
\label{app:segmentation}

\textbf{Gripper events.}
Let $g_1, \ldots, g_n \in \{0, 1\}$ be the gripper channel of a demonstration's actions ($1$ = open). A state change is registered when the new value persists for three consecutive frames, and the event is time-stamped at the first frame of that run. Pairing every close event at frame $c$ with the next open event at frame $r$ (or $r = n$ if the gripper stays closed) gives candidate grasps, of which we keep only those held long enough to lift and safely transport an object to its target location:
\begin{equation}
\mathcal{G} \;=\; \bigl\{ (c_k, r_k) \;:\; r_k - c_k \geq 30 \bigr\}, \qquad c_1 < r_1 \leq c_2 < r_2 \leq \cdots,
\label{eq:app-grasps}
\end{equation}
so brief closures from regrasp adjustments never create spurious stages in the plan.

\textbf{Stage boundaries.}
The gripper-consuming clauses of the plan (pick-and-place and \texttt{turn on}/\texttt{turn off}) are matched in plan order, one-to-one, to the kept grasps in the set $\mathcal{G}$. A cursor $s \gets 1$ then sweeps the plan in a single pass and emits the start frame $s^{(m)}$ of every plan stage:
\begin{equation}
\begin{aligned}
&\texttt{put}(o, \rho, g) \text{ with grasp } (c_k, r_k): && s^{(\texttt{pick})} = s,\;\; s^{(\texttt{place})} = c_k,\;\; s \gets r_k; \\
&\texttt{turn on} / \texttt{turn off} \text{ with grasp } (c_k, r_k): && s^{(\texttt{turn})} = s,\;\; s \gets r_k; \\
&u \in \mathcal{U}_{\mathrm{f}}: && s^{(u)} = s,
\end{aligned}
\label{eq:app-bounds}
\end{equation}
and stage $m$ covers frames $[s^{(m)}, s^{(m+1)})$, the last stage ending at frame $n$. A pick segment therefore ends exactly when the grasp closes, a place segment when it opens, and a gripper-free primitive that follows the last grasp starts at the final release. Every LIBERO-Long training demonstration is segmented by this rule; a demonstration whose grasps and gripper-consuming clauses do not match one-to-one enters training with its original instruction, so all demonstrations contribute to imitation.

\begin{figure}[t]
\centering
\includegraphics[width=\textwidth]{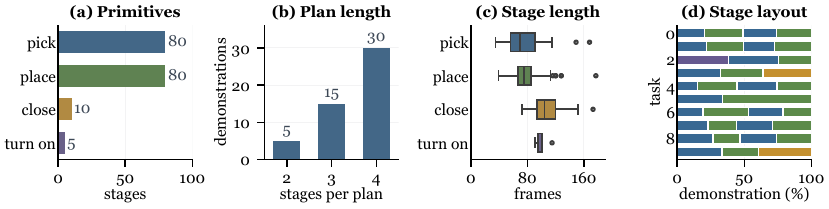}
\caption{Automatic segmentation of the 50 LIBERO-Long demonstrations: (a) stages per primitive; (b) plan length per demonstration; (c) stage length in frames; (d) mean stage layout of every task.}
\label{fig:app_plans}
\end{figure}

\textbf{Supervision from the segments.} Frame $i$ of stage $m$ is paired with $\tilde{x}^{(m)}$ for the Solver and yields stay/advance examples for the pointer. The dwell feature under pointer $j$ is $\lfloor (i - s^{(j)}) / H \rfloor$, the number of chunks since the queried stage began in the demonstration. All 50 LIBERO-Long training demonstrations are segmented into the stages of their plans without a manual label (Figure~\ref{fig:app_plans}).

\section{Proofs}
\label{app:proofs}

\textbf{Conventions.} Decisions $t = 1, \ldots, T$ start at pointer $m_0 = 1$, $d_{\min} \geq 1$, and the dwell obeys
\begin{equation}
d_1 = 0, \qquad d_{t+1} \;=\; \bigl(d_t + 1\bigr)\, \mathbf{1}\bigl[m_t = m_{t-1}\bigr],
\label{eq:app-dwell}
\end{equation}
so it counts the decisions since the last advance. We write $\Delta_t = m_t - m_{t-1}$ for the pointer increment, $t_m = \min\{t : m_t = m\}$ for the first decision of slot $m$ (so $t_1 = 1$, as $d_1 < d_{\min}$), and $\mathcal{S}_m = \{t : m_t = m\}$ for the segment of slot $m$. The chunk of decision $t$ is computed after the pointer update, from $\tilde{x}_t = \tilde{x}^{(m_t)}$. The history $\mathcal{H}_t$ holds $x$, the observations up to $t$ and the earlier chunks.

\subsection{Proof of Proposition~\ref{prop:ns_consistency}}
\label{proof1}

The statement is split into four claims: the pointer is a monotone unit-step counter, its level sets are contiguous segments in plan order, every non-final segment respects the minimum dwell, and every exit is certified by the gate. None of the arguments uses the learned values of $\phi$ or $\theta$.

\begin{lemma}[Binary gate and increment]
\label{lem:gate-binary}
For every $t$ and $u \in \mathcal{U}$, $\Gamma_t(u) \in \{0, 1\}$ and $\Delta_t \in \{0, 1\}$.
\end{lemma}
\begin{proof}
The classes $\mathcal{U}_{\mathrm{g}}, \mathcal{U}_{\mathrm{r}}, \mathcal{U}_{\mathrm{f}}$ partition $\mathcal{U}$, so exactly one class indicator of the gate equals one and
\begin{equation}
\Gamma_t(u) \;\in\; \Bigl\{\, 1,\;\; h_t,\;\; (1 - \kappa_t) \max_{t_{m_{t-1}} \leq t' \leq t} h_{t'} \,\Bigr\} \;\subseteq\; \{0, 1\},
\label{eq:app-gate-values}
\end{equation}
because $\kappa_t \in \{0,1\}$ and $h_{t'} = \kappa_{t'}\, \mathbf{1}[w_{t'} > w_{\min}] \in \{0, 1\}$, and products and maxima of $\{0,1\}$-valued quantities stay in $\{0, 1\}$. By the pointer update, $\Delta_t$ is the product of the four $\{0,1\}$-valued factors $\mathbf{1}[m_{t-1} < M]$, $\mathbf{1}[d_t \geq d_{\min}]$, $\Gamma_t(u^{(m_{t-1})})$ and $\mathbf{1}[\sigma(g_\phi(z_t)) > \frac{1}{2}]$, hence $\Delta_t \in \{0, 1\}$.
\end{proof}

\textbf{Claim 1: monotone unit-step counter.} By Lemma~\ref{lem:gate-binary}, $m_{t-1} \leq m_t \leq m_{t-1} + 1$. If $m_{t-1} = M$, the factor $\mathbf{1}[m_{t-1} < M]$ vanishes; otherwise $m_t \leq m_{t-1} + 1 \leq M$. Induction from $m_0 = 1$ gives
\begin{equation}
1 \;\leq\; m_0 \;\leq\; m_1 \;\leq\; \cdots \;\leq\; m_T \;\leq\; M, \qquad m_t - m_{t-1} \in \{0, 1\} \quad \text{for all } t.
\label{eq:app-monotone}
\end{equation}

\textbf{Claim 2: contiguous segments in plan order.} Let $t < t' < t''$ with $t, t'' \in \mathcal{S}_m$. Monotonicity gives $m = m_t \leq m_{t'} \leq m_{t''} = m$, so $t' \in \mathcal{S}_m$ and every $\mathcal{S}_m$ is an interval of decisions. If $t \in \mathcal{S}_m$ and $t' \in \mathcal{S}_{m'}$ with $m < m'$, then $t \geq t'$ would imply $m = m_t \geq m_{t'} = m'$, a contradiction, so $t < t'$ and the intervals appear in plan order. Unit steps visit every slot $1, \ldots, m_T$ in turn, and
\begin{equation}
\{1, \ldots, T\} \;=\; \bigsqcup_{m=1}^{m_T} \mathcal{S}_m, \qquad \mathcal{S}_m = \bigl[\, t_m,\; t_{m+1} - 1 \,\bigr] \;\; (m < m_T), \qquad \mathcal{S}_{m_T} = \bigl[\, t_{m_T},\; T \,\bigr],
\label{eq:app-partition}
\end{equation}
a partition into $m_T \leq M$ nonempty contiguous segments. Each chunk of $\mathcal{S}_m$ uses $\tilde{x}^{(m)}$, as advances re-predict with the new pointer, so segment $m$ runs under sub-instruction $\mathrm{text}(u^{(m)})$.

\textbf{Claim 3: minimum dwell.} Take a non-final segment, $m < m_T$; its exit at $t_{m+1}$ has $\Delta_{t_{m+1}} = 1$ and therefore $d_{t_{m+1}} \geq d_{\min}$. Unrolling the dwell recursion over the decisions of the segment gives
\begin{equation}
d_t \;=\; t - t_m - \mathbf{1}[m \geq 2] \quad \bigl(t_m < t \leq t_{m+1}\bigr), \qquad \bigl|\mathcal{S}_m\bigr| \;=\; d_{t_{m+1}} + \mathbf{1}[m \geq 2],
\label{eq:app-dwell-length}
\end{equation}
since the advance at $t_m$ resets the dwell at $t_m + 1$. For every $m < m_T$, $|\mathcal{S}_m| \geq d_{t_{m+1}} \geq d_{\min}$, i.e.\ every non-final segment spans at least $d_{\min}$ decisions and at least $H d_{\min}$ environment steps. With $d_{\min} = 2$ and $H = 8$, every non-final segment thus lasts at least 16 environment steps.

\textbf{Claim 4: certified exits.} The exit at $t_{m+1}$ requires $\Gamma_{t_{m+1}}(u^{(m)}) = 1$, which reads by class as
\begin{equation}
u^{(m)} \in \mathcal{U}_{\mathrm{g}} \;\Rightarrow\; h_{t_{m+1}} = 1, \quad
u^{(m)} \in \mathcal{U}_{\mathrm{r}} \;\Rightarrow\; \kappa_{t_{m+1}} = 0 \wedge \exists\, t' \in \bigl[t_m, t_{m+1}\bigr] : h_{t'} = 1,
\label{eq:app-exits}
\end{equation}
so a \texttt{pick} slot is left only while the gripper is commanded closed with the fingers held apart by an object, and a grasp--release slot only after an object was held within the slot and the gripper was reopened. A gripper-free slot is certified by the classifier and the dwell. A \texttt{pick} slot thus never exits on a close command alone, since fingers closing on air fail the width test. Claims 1--4 hold for every rollout and every parameter value of both factors, which proves the full proposition. \qed

\subsection{Proof of Proposition~\ref{prop:graceful_equiv}}
\label{proof2}

\textbf{The high-level factor is a point mass.} With $M = 1$ the factor $\mathbf{1}[m_{t-1} < 1]$ of the pointer update vanishes for every $t$, so by induction $m_t = 1$, independently of $\phi$, the gate and the features $z_t$. Hence
\begin{equation}
\pi^{\mathrm{h}}_\phi\bigl(u \mid \mathcal{H}_t\bigr) \;=\; \mathbf{1}\bigl[u = u^{(1)}\bigr] \qquad \text{for every } \phi \text{ and every history } \mathcal{H}_t.
\label{eq:app-pointmass}
\end{equation}

\textbf{The low-level factor is the backbone.} For a one-slot plan the bridged input is the instruction, $\tilde{x}^{(1)} = x$, so the Solver is the backbone's Laplace chunk policy and the hierarchical policy splits as
\begin{equation}
\pi_\Theta\bigl(u, A_t \mid \mathcal{H}_t\bigr) \;=\; \mathbf{1}\bigl[u = u^{(1)}\bigr]\; \pi_\theta\bigl(A_t \mid I_t, S_t, x\bigr),
\label{eq:app-marginal}
\end{equation}
so, summed over $u$, the chunk marginal of NS-VLA is exactly the backbone policy at every history.

\textbf{Trajectory laws coincide.} Let $\mathcal{K}(s_{t+1}, o_{t+1} \mid s_t, \hat{A}_t)$ be the $H$-step kernel of open-loop execution of the chunk $\hat{A}_t$ from state $s_t$, and $\mu_0$ the initial-state law. Any chunk policy $\pi$ induces the rollout law
\begin{equation}
\mathbb{P}_\pi\bigl(s_{1:T+1}, o_{1:T+1}, \hat{A}_{1:T}\bigr) \;=\; \mu_0(s_1, o_1) \prod_{t=1}^{T} \pi\bigl(\hat{A}_t \mid \mathcal{H}_t\bigr)\, \mathcal{K}\bigl(s_{t+1}, o_{t+1} \mid s_t, \hat{A}_t\bigr).
\label{eq:app-traj}
\end{equation}
NS-VLA and its backbone share $\mu_0$ and $\mathcal{K}$ and place the same conditional law on every chunk, including the exploration noise of the online stage, added to the predicted chunk in the same way. Induction over $t$ shows that the joint laws of $(s_{\leq t}, o_{\leq t}, \hat{A}_{<t})$ coincide for every $t$; hence the trajectory laws coincide, and so do the law of the binary return $R(\tau)$ and every success rate.

\textbf{Training objectives coincide.} A one-slot plan has a single segment and $\tilde{x}^{(1)} = x$, so $x^{\xi}_i = x$ for both values of $\xi$, and in its cross-task decomposition every bridged loss equals the native one:
\begin{equation}
\mathcal{L}_{\mathrm{SFT}}(\theta) \;=\; \tfrac{1}{2}\, \mathcal{L}_{\mathrm{nat}}(\theta) + \tfrac{1}{2} \sum_{u} \lambda_u\, \mathcal{L}_u(\theta) \;=\; \mathbb{E}_{(x,\, o_i,\, A^\star_i) \sim \mathcal{D}}\, \bigl\| f_\theta\bigl(I_i, S_i, x\bigr) - A^\star_i \bigr\|_1,
\label{eq:app-sft-equal}
\end{equation}
the backbone's imitation loss. The pointer is never queried, so the pointer loss plays no role. In the online stage every tuple of $\mathcal{D}^{+}$ carries the input $x$ and, by the previous step, is drawn from the backbone's own trajectory law, so the HJPO objective is the backbone's filtered regression with the same demonstration anchor, and validation selection compares the very same candidates. \qed

\textbf{Derivation of cross-task sharing.} Let $\nu$ be the uniform law on frames and $\mathrm{type}(i)$ the primitive of frame $i$ ($x$ if one-slot), so $\lambda_u = \nu(\mathrm{type} = u)$. As $\xi$ is independent, the tower rule gives
\begin{equation}
\begin{aligned}
\mathcal{L}_{\mathrm{SFT}}(\theta) &\;=\; \tfrac{1}{2}\, \mathbb{E}_\nu\bigl[\ell_\theta(i; x)\bigr] + \tfrac{1}{2}\, \mathbb{E}_\nu\bigl[\ell_\theta(i; \tilde{x}^{(m)})\bigr] \\
&\;=\; \tfrac{1}{2}\, \mathcal{L}_{\mathrm{nat}}(\theta) + \tfrac{1}{2} \sum_{u} \lambda_u\, \mathbb{E}_\nu\bigl[\ell_\theta(i; \tilde{x}^{(m)}) \bigm| \mathrm{type}(i) = u\bigr],
\end{aligned}
\label{eq:app-share}
\end{equation}
with $\ell_\theta(i; x') = \|f_\theta(I_i, S_i, x') - A^\star_i\|_1$. The conditional expectation is $\mathcal{L}_u(\theta)$, the mean bridged loss over all frames of primitive $u$, pooled over tasks, which proves the decomposition.

\subsection{Proof of Proposition~\ref{prop:rl_convergence}}
\label{proof3}

\textbf{Factorized likelihood.} Given the stage features as fixed inputs, the pointer is a deterministic function of the stay/advance decisions $y_{1:T}$ and the gate, and the primitive executed at decision $t$ is $u^{(m_t)}$. Writing the likelihood of a rollout with the decision $y_t \sim p_\phi(\cdot \mid z_t)$ made before the chunk,
\begin{equation}
\pi_\Theta(\tau) \;=\; \mu_0(s_1, o_1) \prod_{t=1}^{T} p_\phi\bigl(y_t \mid z_t\bigr)\, \pi^{\mathrm{l}}_\theta\bigl(\hat{A}_t \mid u^{(m_t)}, \mathcal{H}_t\bigr)\, \mathcal{K}\bigl(s_{t+1}, o_{t+1} \mid s_t, \hat{A}_t\bigr),
\label{eq:app-lik}
\end{equation}
and taking logarithms leaves an environment term $\log \mu_{\mathrm{env}}(\tau)$, the sum of $\log \mu_0$ and the kernel terms $\log \mathcal{K}$, that depends on neither $\phi$ nor $\theta$. The gradient in $\phi$ thus involves only the decision terms $\log p_\phi(y_t \mid z_t)$ and the gradient in $\theta$ only the chunk terms $\log \pi^{\mathrm{l}}_\theta(\hat{A}_t \mid u^{(m_t)}, \mathcal{H}_t)$, so the factors get disjoint gradients. HJPO holds $\phi$ fixed online and optimizes the second sum only.

\textbf{Setup.} Fix an iteration and let $\mathcal{C} = \mathcal{D}^{+} \cup \mathcal{D}^{-}$ be the collected chunks, where every chunk $c$ inherits the binary reward $R(c) = R(\tau(c))$ of its rollout, $\mathcal{D}^{+}$ holds the $N^{+} \geq 1$ chunks with $R = 1$ and $\mathcal{D}^{-}$ the $N^{-}$ chunks with $R = 0$. Write $\ell_\theta(c) = \|f_\theta(I_c, S_c, \tilde{x}_c) - \hat{A}_c\|_1$ and $\varepsilon = e^{-1/\beta}$.

\textbf{AWR reduces to weighted L1 regression.} AWR~\citep{Advantage-Weighted-Regression} fits the policy by weighted maximum likelihood with the AWR weights $w_\beta$ of each chunk, self-normalized over the collected data, $\bar{w}_\beta(c) = w_\beta(c) / \sum_{c' \in \mathcal{C}} w_\beta(c')$. Self-normalization removes any constant baseline, since for every $b_0 \in \mathbb{R}$
\begin{equation}
\bar{w}_\beta(c) \;=\; \frac{\exp\bigl((R(c) - b_0)/\beta\bigr)}{\sum_{c' \in \mathcal{C}} \exp\bigl((R(c') - b_0)/\beta\bigr)} \;=\; \frac{\exp\bigl(R(c)/\beta\bigr)}{\sum_{c' \in \mathcal{C}} \exp\bigl(R(c')/\beta\bigr)},
\label{eq:app-baseline}
\end{equation}
so the choice $R_{\max}$ only fixes the scale of the unnormalized weights. Under the Laplace chunk policy, $-\log \pi^{\mathrm{l}}_\theta(\hat{A}_c \mid \cdot) = \ell_\theta(c)/b + 7H \log(2b)$, whose constant does not depend on $\theta$; therefore
\begin{equation}
b\, \mathcal{L}^{\mathrm{AWR}}_\beta(\theta) \;=\; -\,b \sum_{c \in \mathcal{C}} \bar{w}_\beta(c) \log \pi^{\mathrm{l}}_\theta\bigl(\hat{A}_c \mid \cdot\bigr) \;=\; \sum_{c \in \mathcal{C}} \bar{w}_\beta(c)\, \ell_\theta(c) \;+\; 7Hb \log(2b).
\label{eq:proof3-awr}
\end{equation}

\textbf{Zero-temperature limit with explicit rate.} With $R_{\max} = 1$, $w_\beta = 1$ on $\mathcal{D}^{+}$ and $w_\beta = \varepsilon$ on $\mathcal{D}^{-}$, so $\bar{w}_\beta(c) = 1/(N^{+} + N^{-}\varepsilon)$ on $\mathcal{D}^{+}$ and $\varepsilon/(N^{+} + N^{-}\varepsilon)$ on $\mathcal{D}^{-}$. Let $L^{+}(\theta) = \frac{1}{N^{+}} \sum_{c \in \mathcal{D}^{+}} \ell_\theta(c)$ be the replay term of $\mathcal{L}_{\mathrm{HJPO}}$ under uniform sampling from $\mathcal{D}^{+}$ and $L^{-}(\theta)$ the mean over $\mathcal{D}^{-}$. Then
\begin{equation}
\begin{aligned}
\sum_{c \in \mathcal{C}} \bar{w}_\beta(c)\, \ell_\theta(c) - L^{+}(\theta) &\;=\; \frac{N^{-}\varepsilon}{N^{+} + N^{-}\varepsilon}\, \bigl(L^{-}(\theta) - L^{+}(\theta)\bigr), \\
\Bigl|\sum_{c \in \mathcal{C}} \bar{w}_\beta(c)\, \ell_\theta(c) - L^{+}(\theta)\Bigr| &\;\leq\; \frac{N^{-}}{N^{+}}\, e^{-1/\beta}\, \bigl|L^{-}(\theta) - L^{+}(\theta)\bigr|.
\end{aligned}
\label{eq:proof3-rate}
\end{equation}
The right-hand side vanishes as $\beta \to 0$ at rate $e^{-1/\beta}$, pointwise in $\theta$ and uniformly on any set where the losses are bounded. The identity is linear in $\ell_\theta$, so the same bound holds for the (sub)gradients,
\begin{equation}
\Bigl\| \nabla_\theta \sum_{c \in \mathcal{C}} \bar{w}_\beta(c)\, \ell_\theta(c) - \nabla_\theta L^{+}(\theta) \Bigr\| \;\leq\; \frac{N^{-}}{N^{+}}\, e^{-1/\beta}\, \bigl\| \nabla_\theta L^{-}(\theta) - \nabla_\theta L^{+}(\theta) \bigr\|,
\label{eq:proof3-grad}
\end{equation}
so the AWR update direction converges to the gradient of the replay term. As the demonstration term is free of $\beta$, the anchored objective converges to $\mathcal{L}_{\mathrm{HJPO}}$ up to the constant $(1 - \eta)\, 7Hb \log(2b)$:
\begin{equation}
\begin{aligned}
&\Bigl| \eta\, \mathcal{L}_{\mathrm{SFT}}(\theta) + (1 - \eta)\bigl(b\, \mathcal{L}^{\mathrm{AWR}}_\beta(\theta) - 7Hb \log(2b)\bigr) - \mathcal{L}_{\mathrm{HJPO}}(\theta) \Bigr| \\
&\qquad\;\leq\; (1 - \eta)\, \frac{N^{-}}{N^{+}}\, e^{-1/\beta}\, \bigl|L^{-}(\theta) - L^{+}(\theta)\bigr|.
\end{aligned}
\label{eq:proof3-hjpo}
\end{equation}
An iteration with an empty $\mathcal{D}^{+}$ performs no replay update and keeps the current $\theta$ unchanged.

\textbf{Target law.} Let $\mu$ be the trajectory law of the exploring hierarchical policy and $q = \mu(R = 1) > 0$. The AWR target is the reweighted law $\mu_\beta(\tau) = \mu(\tau)\, w_\beta(\tau) / Z_\beta$ with $Z_\beta = q + (1 - q)\varepsilon$, and the success-conditioned law is $\mu^{+}(\tau) = \mu(\tau)\, \mathbf{1}[R(\tau) = 1] / q$. On successes $\mu_\beta - \mu^{+} = \mu\,(1/Z_\beta - 1/q) \leq 0$ and on failures $\mu_\beta - \mu^{+} = \mu\, \varepsilon / Z_\beta \geq 0$, so the total variation distance is
\begin{equation}
\mathrm{TV}\bigl(\mu_\beta, \mu^{+}\bigr) \;=\; \tfrac{1}{2}\Bigl[\, q\Bigl(\tfrac{1}{q} - \tfrac{1}{Z_\beta}\Bigr) + \tfrac{(1 - q)\varepsilon}{Z_\beta} \Bigr] \;=\; \frac{(1 - q)\, e^{-1/\beta}}{q + (1 - q)\, e^{-1/\beta}} \;\leq\; \frac{1 - q}{q}\, e^{-1/\beta},
\label{eq:proof3-tv}
\end{equation}
and the Kullback--Leibler divergence $\mathrm{KL}(\mu^{+} \,\|\, \mu_\beta) = \log(Z_\beta / q) \leq \frac{1-q}{q} e^{-1/\beta}$ obeys the same rate. As $\phi$ and $z_t$ are fixed and the environment terms are free of $\theta$, maximum likelihood on $\mu^{+}$ satisfies
\begin{equation}
\argmax_\theta\; \mathbb{E}_{\tau \sim \mu^{+}}\bigl[\log \pi_\Theta(\tau)\bigr] \;=\; \argmin_\theta\; \mathbb{E}_{\tau \sim \mu^{+}}\Bigl[\sum_{t=1}^{T} \ell_\theta(c_t)\Bigr], \qquad c_t = \bigl(o_t, \tilde{x}_t, \hat{A}_t\bigr),
\label{eq:proof3-mle}
\end{equation}
whose empirical counterpart is the replay term $L^{+}$: filtered AWR fits the policy's own successes.

\textbf{Validation selection is non-degrading.} The reported model is $\hat{\theta} = \theta_{\hat{k}}$ with $\hat{k} = \argmax_{k \in \mathcal{Q}} \mathrm{SR}_{\mathrm{val}}(\theta_k)$ over the candidate iterates $\mathcal{Q}$, and $0 \in \mathcal{Q}$ indexes the supervised start. Hence
\begin{equation}
\mathrm{SR}_{\mathrm{val}}(\hat{\theta}) \;=\; \max_{k \in \mathcal{Q}} \mathrm{SR}_{\mathrm{val}}(\theta_k) \;\geq\; \mathrm{SR}_{\mathrm{val}}(\theta_0),
\label{eq:proof3-select}
\end{equation}
and tie-breaking chooses only among maximizers, leaving the value unchanged. The inequality holds deterministically for every online run, as all candidates are scored on the same validation states. \qed

\section{Training algorithm}
\label{sec:rl_details}

Algorithm~\ref{alg:nsgrpo} gives the full training procedure and Algorithm~\ref{alg:rollout} a rollout of the hierarchical policy. Following the options framework~\citep{Between-MDPs-and-semi-MDPs}, a \emph{decision} $t$ updates the pointer and emits a chunk, which then runs open-loop for $H = 8$ environment steps before the next decision reads a new observation.

\label{sec:rl_algo}
\begin{algorithm}[H]
\caption{NS-VLA training: supervised stage, online HJPO stage and validation selection}
\label{alg:nsgrpo}
\begin{algorithmic}[1]
\Require demonstrations $\mathcal{D}$ (5 per task); parser $\mathcal{P}$; backbone $f_\theta$ initialized from the generic vision-language model; validation initial states $\mathcal{V}$ (5 per task); $K = 8$, $N_{\mathrm{roll}} = 8$, $U = 400$, $\eta = 0.5$, $\sigma_{\mathrm{exp}} = 0.03$
\Statex \textbf{Stage I: supervised fitting of both factors}
\For{each demonstration $(x, o_{1:n}, a_{1:n}) \in \mathcal{D}$}
  \State $p \gets \mathcal{P}(x)$ \Comment{plan grammar}
  \State $(s^{(1)}, \ldots, s^{(M)}) \gets$ gripper-event segmentation of $a_{1:n}$ \Comment{grasps and stage starts}
\EndFor
\State train $\theta$ on $\mathcal{L}_{\mathrm{SFT}}$ \Comment{$\xi \sim \mathrm{Bern}(1/2)$}
\State $\theta_0 \gets$ snapshot with the highest $\mathrm{SR}_{\mathrm{val}}$ \Comment{snapshots every 2.5k steps}
\State build stay/advance examples with $f_{\theta_0}$ under the queried pointer; fit $\phi$ on $\mathcal{L}_{\mathrm{ptr}}$ \Comment{class-balanced}
\Statex \textbf{Stage II: online stage of the low-level factor (filtered AWR)}
\State $\theta \gets \theta_0$; \ $\mathcal{D}^{+} \gets \emptyset$; \ $\mathcal{Q} \gets \{0\}$
\For{$k = 1, \ldots, K$}
  \For{each task, $N_{\mathrm{roll}}$ times}
    \State $(\tau, R) \gets \textsc{Rollout}(\theta, \phi, \sigma_{\mathrm{exp}})$ \Comment{Algorithm~\ref{alg:rollout}}
    \If{$R = 1$} $\mathcal{D}^{+} \gets \mathcal{D}^{+} \cup \{(o_t, \tilde{x}_t, \hat{A}_t)\}_{t \in \tau}$ \Comment{task-level credit}
    \EndIf
  \EndFor
  \If{$\mathcal{D}^{+} \neq \emptyset$}
    \State take $U$ AdamW steps on $\mathcal{L}_{\mathrm{HJPO}}(\theta)$ from the current $\theta$ \Comment{replay mixed with demos}
  \EndIf
  \State $\theta_k \gets \theta$
  \If{$k \bmod 2 = 0$}
    \State evaluate the policy with $\theta_k$ on $\mathcal{V}$; \ $\mathcal{Q} \gets \mathcal{Q} \cup \{k\}$
  \EndIf
\EndFor
\State \Return $\hat{\theta} \gets \theta_{\hat{k}}$, \ $\hat{k} = \argmax_{k \in \mathcal{Q}} \mathrm{SR}_{\mathrm{val}}(\theta_k)$ \Comment{ties: more milestones, then larger $k$}
\end{algorithmic}
\end{algorithm}

\begin{algorithm}[t]
\caption{\textsc{Rollout} of one episode with the hierarchical policy (inference uses $\sigma_{\mathrm{exp}} = 0$)}
\label{alg:rollout}
\begin{algorithmic}[1]
\Require parameters $(\theta, \phi)$; exploration scale $\sigma_{\mathrm{exp}}$; $d_{\min} = 2$; $w_{\min} = 0.0025$
\State reset the environment; read $x$; $p \gets \mathcal{P}(x)$ \Comment{plan, once per episode}
\State $m \gets 1$; $d \gets 0$; $\textit{held} \gets 0$; $\kappa \gets 0$; $h \gets 0$ \Comment{pointer, dwell and gripper state}
\While{the episode is not over}
  \State observe $o_t = (I_t, S_t)$; \ $\tilde{x} \gets$ bridged input of slot $m$ ($x$ itself if $M = 1$)
  \State $(\bar{A}_t, \bar{h}_t) \gets f_\theta(I_t, S_t, \tilde{x})$ \Comment{mode chunk and pooled context, one pass}
  \If{$M > 1$}
    \State $z_t \gets [\bar{h}_t;\, e(u^{(m)});\, e(u^{(\min(m+1, M))});\, S_t;\, d;\, m/M]$ \Comment{stage features}
    \State $\Gamma \gets$ gate of $u^{(m)}$ from $(\kappa, h, \textit{held})$ \Comment{precondition gate}
    \If{$m < M$ \textbf{and} $d \geq d_{\min}$ \textbf{and} $\Gamma = 1$ \textbf{and} $\sigma(g_\phi(z_t)) > 1/2$}
      \State $m \gets m + 1$; \ $d \gets 0$; \ $\textit{held} \gets h$ \Comment{a grasp held across the boundary counts}
      \State $\tilde{x} \gets$ bridged input of slot $m$; \ $(\bar{A}_t, \cdot) \gets f_\theta(I_t, S_t, \tilde{x})$ \Comment{re-predict}
    \Else
      \State $d \gets d + 1$
    \EndIf
  \EndIf
  \State $\hat{A}_t \gets \bar{A}_t$; \ $\hat{A}_t[:, 1{:}6] \gets \mathrm{clip}\bigl(\bar{A}_t[:, 1{:}6] + \zeta, -1, 1\bigr)$, $\zeta \sim \mathcal{N}(0, \sigma_{\mathrm{exp}}^2 I)$ \Comment{arm only}
  \State execute $\hat{A}_t$ open-loop for up to $H$ steps (stop at success)
  \State $\kappa \gets$ last executed gripper command; \ $w \gets$ finger separation from $S$
  \State $h \gets \kappa\, \mathbf{1}[w > w_{\min}]$; \ $\textit{held} \gets \max(\textit{held}, h)$
  \State record $(o_t, \tilde{x}, \hat{A}_t)$ with the gripper dimension binarized
\EndWhile
\State \Return trajectory $\tau$ and task reward $R(\tau) \in \{0, 1\}$
\end{algorithmic}
\end{algorithm}

\subsection{Implementation of the two stages}
\label{app:rl_impl_details}

\textbf{Supervised stage.} The Solver is trained on the sub-instruction-augmented loss with snapshots every 2.5k steps, and the snapshot with the highest validation SR becomes $\theta_0$. The stage features $z_t$ are then computed by $\theta_0$ under the bridged input of the queried pointer, exactly as at execution time. Each frame $i$ of segment $j$ yields a stay example $(z^{(j)}_i, 0)$, with $z^{(j)}_i$ its stage features under pointer $j$, and, for $j > 1$, an advance example $(z^{(j-1)}_i, 1)$, and with class weights $\omega_y = N / (2 N_y)$ over $N_y$ examples of class $y$ out of $N$, the classifier $g_\phi$ minimizes the class-balanced cross-entropy
\begin{equation}
\mathcal{L}_{\mathrm{ptr}}(\phi) \;=\; -\,\mathbb{E}_{(z, y)}\Bigl[\omega_1\, y \log \sigma\bigl(g_\phi(z)\bigr) \;+\; \omega_0\, (1 - y) \log\bigl(1 - \sigma(g_\phi(z))\bigr)\Bigr],
\label{eq:ptr}
\end{equation}
which balances the rare advance decisions against the far more frequent stay decisions.

\textbf{Online stage.} A replayed tuple stores both images, the proprioceptive state, the bridged input under which the chunk ran and the executed chunk with a binarized gripper dimension, so the target is the action that produced the success and the condition is the sub-instruction that the pointer selected. $\mathcal{D}^{+}$ accumulates over iterations; each sample comes from the demonstrations with probability $\eta = 0.5$ and otherwise uniformly from $\mathcal{D}^{+}$. Every update starts a fresh AdamW optimizer on the LoRA adapters, the action head and the proprioception projection, while parser, gate and $\phi$ stay fixed.

\textbf{Rollouts and selection.} Rollouts start from training initial states, validation uses five held-out initial states per task, and the official test states are reserved for evaluation, so selection never touches test episodes. Simulator milestones serve analysis and the tie-break of the selection only.

\section{Baselines and ablation variants}
\label{sec:baseline_details}

All ten baselines and VLA-Adapter train one multi-task model per suite on the same 10\% split with seed 7 (VLA-Adapter also 17 and 27 on Long), follow their official recipes from scratch or general-purpose weights, and are evaluated with their final checkpoint (VLA-Adapter, like NS-VLA, with its best-validation snapshot) under the program, initial states, checks and limits of NS-VLA.

\begin{itemize}
    \item \textbf{BC-RNN}~\citep{robomimic}: a recurrent behavior-cloning policy over a ten-step observation history with a five-mode Gaussian-mixture action head, ResNet image encoders and FiLM language fusion of a BERT instruction embedding.
    \item \textbf{BC-Transformer (BC-T)}~\citep{LIBERO}: the LIBERO multi-task policy that replaces the recurrent core of BC-RNN by a temporal Transformer over the same observation history and Gaussian-mixture head.
    \item \textbf{BC-ViLT}~\citep{LIBERO}: the LIBERO vision-language Transformer that tokenizes image patches and the instruction embedding jointly in a single Transformer.
    \item \textbf{ACT}~\citep{Learning-Fine-Grained-Bimanual-Manipulation-with-Low-Cost-Hardware}: a conditional-VAE Transformer that predicts 16-step action chunks.
    \item \textbf{Diffusion Policy (DP)}~\citep{Diffusion-policy}: a denoising-diffusion policy over 32-step action chunks, with 100 training and 10 DDIM inference steps.
    \item \textbf{VQ-BeT}~\citep{VQ-BeT}: a behavior Transformer over residual vector-quantized action codes of five-step blocks with a ten-frame observation stack.
    \item \textbf{QueST}~\citep{QueST}: a two-stage policy that learns a finite-scalar-quantized skill autoencoder over 32-step action blocks and an autoregressive GPT prior over the skill tokens.
    \item \textbf{BAKU}~\citep{BAKU}: a compact multi-task Transformer with ResNet encoders, FiLM conditioning on a MiniLM sentence embedding of the instruction, proprioceptive input and ten-step action chunks with temporal aggregation.
    \item \textbf{Octo}~\citep{Octo}: the 93M-parameter Octo-Base generalist Transformer policy with a T5-base instruction encoder and a diffusion action head, fine-tuned from octo-base-1.5 with its official recipe.
    \item \textbf{SmolVLA}~\citep{SmolVLA}: a 0.45B vision-language-action model that pairs the SmolVLM2-500M vision-language model with a flow-matching action expert over 50-step chunks, fine-tuned from smolvla\_base.
    \item \textbf{VLA-Adapter}~\citep{VLA-Adapter}: the backbone of NS-VLA, initialized from the same generic vision-language model and trained with the same supervised recipe on the original instruction.
\end{itemize}

\textbf{Codebases.} BC-RNN, BC-T and BC-ViLT use the official LIBERO codebase (commit \texttt{8f1084e}, with the end-effector input set to the 6-D pose of our state). ACT, DP, VQ-BeT and QueST use the QueST codebase (commit \texttt{cc25a83}) with CLIP instruction embeddings. BAKU uses its official repository unmodified (commit \texttt{47e3670}), Octo the official repository unmodified (commit \texttt{241fb35}) with its fine-tuning script, and SmolVLA the LeRobot package (version 0.4.4). Observations are the third-person and wrist images (128$\times$128 for the LIBERO, QueST and BAKU codebases; 256$\times$256 and 128$\times$128 for Octo) together with the end-effector and gripper state, and actions are the same 7-D end-effector deltas with gripper command used by NS-VLA. Table~\ref{tab:baseline_budget} lists every training configuration.

\begin{table}[!htb]
\centering
\caption{Training configuration; lengths per suite (S/O/G/L), ep.: epochs, AE: autoencoder.}
\label{tab:baseline_budget}
\vspace{2pt}
\fontsize{8pt}{10pt}\selectfont
\setlength{\tabcolsep}{3.6pt}
\begin{tabular}{@{}llrrll@{}}
\toprule
\textbf{Method} & \textbf{Initialization} & \textbf{Params} & \textbf{Batch} & \textbf{Learning rate} & \textbf{Length (S / O / G / L)} \\
\midrule
BC-RNN & from scratch & 19.10M & 32 & 1e-4, cosine & 50 / 50 / 50 / 28 ep. \\
BC-T & from scratch & 5.38M & 32 & 1e-4, cosine & 50 / 50 / 50 / 21 ep. \\
BC-ViLT & from scratch & 7.90M & 32 & 1e-4, cosine & 30 / 30 / 30 / 15 ep. \\
ACT & from scratch & -- & 128 & 1e-4 & 100 / 100 / 100 / 100 ep. \\
DP & from scratch & 49.88M & 128 & 1e-4 & 200 / 197 / 200 / 173 ep. \\
VQ-BeT & from scratch & 1.10M & 128 & 5.5e-5 & 100 / 99 / 61 / 44 ep. (+100 VQ-VAE) \\
QueST & from scratch & -- & 128 & 1e-4 & 100 / 100 / 100 / 100 ep. (+100 AE) \\
BAKU & from scratch & 6.53M & 64 & 1e-4 & 28.3k / 32.8k / 36.6k / 35.4k steps \\
Octo & octo-base-1.5 & 93M & 48 & 3e-4, cosine & 2.6k / 2.6k / 2.6k / 2.6k steps \\
SmolVLA & smolvla\_base & 0.45B & 16 & 1e-4, cosine & 4.5k / 4.5k / 4.0k / 4.0k steps \\
\rowcolor{nslight}
\shadeL VLA-Adapter & generic VLM & $\sim$1.3B & 16 & 2e-4, cosine & 10k / 10k / 10k / 20k steps\shadeR \\
\bottomrule
\end{tabular}
\end{table}

The ten baselines train on NVIDIA V100 GPUs, and VLA-Adapter and NS-VLA on NVIDIA H100 GPUs in BF16; every method is evaluated on the same machines with EGL rendering.

\textbf{Ablation variants.} All variants of Table~\ref{tab:ablation} and Figure~\ref{fig:rq2_ablation} run on the 500 LIBERO-Long test episodes with seed 7. \emph{w/o Encoder} runs the final checkpoint without plan and pointer, so the backbone reads the original instruction. \emph{w/o gate} sets $\Gamma_t \equiv 1$, so the pointer advances on the classifier and the dwell alone. \emph{w/o Solver} is the seed-7 VLA-Adapter run, trained on the original instructions and given the same online stage without plan or bridged input; its supervised start collapses on LIBERO-Long (0.0) and collects no success, so it is kept. \emph{w/o HJPO} evaluates the supervised start $\theta_0$ with the Encoder.

\section{Datasets and evaluation protocol}
\label{sec:dataset_details}

\textbf{LIBERO.} LIBERO~\citep{LIBERO} has four suites of ten tasks with 50 demonstrations and 50 official test initial states per task. Spatial varies the layout around identical objects, Object the target object, Goal the goal in a fixed scene, and Long chains two to four primitives in a single instruction.

\textbf{Training split.} Every method trains on the first five demonstrations of every task in the fixed record order of the no-op-filtered LIBERO RLDS release, 200 demonstrations in total (33,604 frames). The list is shared by all methods and seeds, and normalization statistics use these demonstrations only.

\textbf{LIBERO-Plus.} LIBERO-Plus~\citep{LIBERO-Plus} is held out from training and selection. Its 10,030 variants (2,402 Spatial, 2,518 Object, 2,591 Goal, 2,519 Long) each perturb one of seven dimensions at a difficulty level from 1 to 5; the language dimension rewrites the instruction, as shown in Figure~\ref{fig:rq6_case}.

\textbf{Evaluation protocol.} LIBERO uses the first 20 official initial states of every task, 200 episodes per suite. LIBERO-Plus draws 300 variants per suite, stratified by dimension: 100 with sampling seed 2026 and a pre-registered extension of 200 with seed 2027, 1,200 variants (129 background, 191 camera, 185 language, 136 lighting, 182 layout, 186 robot-state, 191 sensor-noise). Lists are fixed in advance and shared, so every comparison is paired; the extension runs the same checkpoints on a second machine whose renderings differ by at most two gray levels. VLA-Adapter collapses on LIBERO-Long with seed 7 (0.0), so Table~\ref{tab:main_results} reports its mean over seeds 7, 17 and 27 (0.0, 42.5 and 45.5) with the standard error across seeds. LIBERO-Plus uses the seed-7 checkpoints, so VLA-Adapter's Long variants come from this collapsed model (0 of 300); on the 900 Spatial, Object and Goal variants it solves 545 against 542 for NS-VLA, and with seed 17 it solves 31.2\% of all 2,519 Long variants against 47.1\%. A cell with no success in $n$ episodes shows the one-sided 95\% Clopper--Pearson interval $[0, u]$, $u = 1 - 0.05^{1/n}$, as $u/2 \pm u/2$: $0.8 \pm 0.8$ for $n = 186$ or 191 and $1.1 \pm 1.1$ for $n = 129$. RQ2--RQ5 mainly analyze the 500 LIBERO-Long test episodes.

\section{Implementation details}
\label{sec:implementation_details}

Table~\ref{tab:hyperparams} lists the full configuration. The backbone is VLA-Adapter initialized from a generic vision-language model without robot pretraining, with randomly initialized action head and proprioception projection. LIBERO-Long instructions are parsed into the plans of Table~\ref{tab:long_plans}; on Spatial, Object and Goal the parser is not applied, by design, and every episode there runs the one-slot plan, on which NS-VLA reduces exactly to its backbone trained with the same procedure (Proposition~\ref{prop:graceful_equiv}).

\begin{table}[!htb]
\centering
\caption{Hyperparameters of NS-VLA, of its two training stages and of the evaluation protocol.}
\label{tab:hyperparams}
\vspace{2pt}
\fontsize{8pt}{10pt}\selectfont
\setlength{\tabcolsep}{4pt}
\renewcommand{\arraystretch}{1.05}
\begin{tabular}{@{}ll>{\raggedright\arraybackslash}p{0.46\linewidth}@{}}
\toprule
\textbf{Group} & \textbf{Hyperparameter} & \textbf{Value} \\
\midrule
\multirow{5}{*}{Backbone} & Language model / vision encoders & Qwen2.5-0.5B / DINOv2 + SigLIP \\
 & Adaptation & LoRA rank 64 \\
 & Action head & L1 regression, 7-D actions \\
 & Action chunk $H$ & 8 steps, executed open-loop \\
 & Initialization & generic vision-language model \\
\midrule
\multirow{5}{*}{Plan and segments} & Primitive vocabulary & pick, place, open, close, turn on, turn off, push \\
 & Pick-and-place clause & split into pick + place \\
 & Pick segment end & gripper close persisting $\geq$ 30 frames \\
 & Place segment end & next gripper open \\
 & Bridged input & ``\texttt{<instruction>. now: <sub-instruction>}'' \\
\midrule
\multirow{6}{*}{Supervised stage} & Optimizer & AdamW, peak lr $2{\times}10^{-4}$ \\
 & Schedule & 100-step warmup, cosine decay to $2{\times}10^{-5}$ \\
 & Batch size / precision & 16 / BF16 \\
 & Steps & 20k (Long), 10k (Spatial, Object, Goal) \\
 & Sub-instruction probability & 0.5 \\
 & Snapshot selection & every 2.5k steps, validation SR \\
\midrule
\multirow{6}{*}{Stage pointer} & Classifier $g_\phi$ & stay/advance, class-balanced cross-entropy \\
 & Training & AdamW, lr $3{\times}10^{-4}$, 30 epochs \\
 & Advance threshold & $\sigma(g_\phi(z_t)) > 1/2$ \\
 & Minimum dwell $d_{\min}$ & 2 chunks \\
 & Pick gate & closed, finger gap $> w_{\min} = 0.0025$ \\
 & Place gate & object held, then gripper opened within the stage \\
\midrule
\multirow{7}{*}{Online stage (HJPO)} & Iterations $K$ & 8 \\
 & Rollouts & 8 per task per iteration (80 per suite) \\
 & Exploration noise & Gaussian, $\sigma_{\mathrm{exp}} = 0.03$ on the 6 arm dimensions \\
 & Updates per iteration & 400, lr $5{\times}10^{-5}$ \\
 & Batch size & 16 (8 $\times$ 2 accumulation) \\
 & Demonstration ratio $\eta$ & 0.5 \\
 & Updated parameters & LoRA, action head, proprioception projection \\
\midrule
\multirow{2}{*}{Selection} & Validation states & 5 held-out initial states per task \\
 & Candidates & iterations $\{0, 2, 4, 6, 8\}$, best validation SR \\
\midrule
\multirow{4}{*}{Evaluation} & Initial states & official LIBERO test states \\
 & Max episode steps & 220 / 280 / 300 / 520 (S / O / G / L) \\
 & Seed & 7 (also 17 and 27 for Figures~\ref{fig:rq4_solver} and \ref{fig:rq5_rl}(d) and VLA-Adapter on Long) \\
 & Rendering & EGL, the same machines for all methods \\
\bottomrule
\end{tabular}
\end{table}

\section{Experiment details}
\label{app:exp_details}

Three episode sets are used: the 200 episodes per LIBERO suite of Table~\ref{tab:main_results}, the 500 LIBERO-Long test episodes of RQ2--RQ5, and the 1,200 LIBERO-Plus variants, with the paraphrase analysis using all 1,537 language variants. On LIBERO-Long, NS-VLA thus reaches $42.0$ on the 200 episodes, $40.6$ on the 500 test episodes and $45.3$ on average over the three training seeds (Figure~\ref{fig:rq4_solver}).

\subsection{Main results (RQ1)}
\begin{figure}[!htb]
\centering
\includegraphics[width=\textwidth]{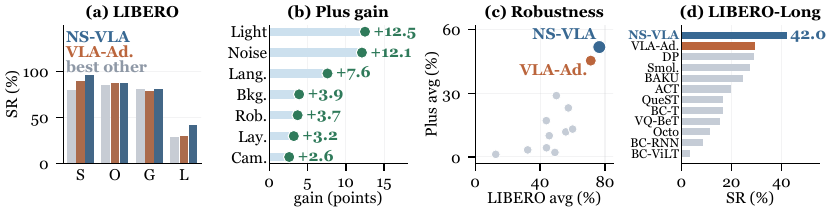}
\caption{Main results: (a) LIBERO suites (S/O/G/L); (b) gain over the best baseline per LIBERO-Plus dimension; (c) LIBERO against LIBERO-Plus average of every method; (d) LIBERO-Long.}
\label{fig:app_results}
\end{figure}

\textbf{Setting.} A cell of Table~\ref{tab:main_results} is an SR with its binomial standard error, the LIBERO average is the mean of the four suites, the LIBERO-Plus average pools the 1,200 variants, and Figure~\ref{fig:app_results} charts them.

\textbf{Results.} NS-VLA ranks first or ties for first in all thirteen columns, leads LIBERO by $5.3$ and LIBERO-Plus by $6.4$ points on average, and is the only method in the upper right of panel (c). It wins $77$ more of the 1,200 variants than VLA-Adapter (exact McNemar test, $p = 1.5\times10^{-5}$).

\textbf{Data and interaction.} All methods share the 10\% split, and NS-VLA additionally interacts online during HJPO. Without its online stage, NS-VLA uses exactly the data of the baselines and no interaction, yet averages $41.2$ on the 500 LIBERO-Long test episodes over seeds 7, 17 and 27 against $29.5$ for VLA-Adapter. Flat RL in the main text uses seed 7, whose start scores 0.0 and collects no success; with seeds 17 and 27 it collects 239 and 227 of 640 (HJPO: 307 and 192).

\subsection{Ablation study (RQ2)}
\textbf{Setting.} Table~\ref{tab:app_ablation} adds milestones per episode, the share of episodes that reach four stages, and the SR change with a 95\% interval from a paired bootstrap over the ten tasks (10,000 resamples, seed 0). Figure~\ref{fig:rq2_ablation}(c) shows the share of episodes reaching stage $k+1$ among those reaching stage $k$.

\begin{table}[!htb]
\centering
\caption{Extended ablations of every component on the 500 LIBERO-Long test episodes (seed 7).}
\label{tab:app_ablation}
\vspace{2pt}
\fontsize{8pt}{10pt}\selectfont
\setlength{\tabcolsep}{5pt}
\begin{tabular}{@{}lccccc@{}}
\toprule
\textbf{Variant} & \textbf{SR} & \textbf{PSCR} & \textbf{Milestones} & \textbf{4 stages (\%)} & \textbf{$\Delta$SR [95\% CI]} \\
\midrule
\rowcolor{nslight}
\shadeL\textbf{NS-VLA} & \textbf{40.6} & \textbf{54.4} & \textbf{1.79} & \textbf{14.2} & --\shadeR \\
w/o gate & 39.0 & 53.9 & 1.77 & 13.0 & $-1.6$ [$-3.6$, $0.0$] \\
w/o Encoder & 38.0 & 53.1 & 1.75 & 12.8 & $-2.6$ [$-8.0$, $+1.2$] \\
w/o HJPO & 28.2 & 48.9 & 1.62 & 8.4 & $-12.4$ [$-19.8$, $-5.0$] \\
w/o Solver & 0.0 & 7.3 & 0.25 & 0.0 & $-40.6$ [$-57.0$, $-24.4$] \\
\bottomrule
\end{tabular}
\end{table}

\textbf{Results.} The full model leads every variant on all four measures. Its continuation rates are $77.0$, $63.0$ and $40.1\%$ up to the fourth stage, against $67.0$, $51.2$ and $32.1\%$ without HJPO refinement.

\subsection{Neuro-Symbolic Encoder (RQ3)}
\textbf{Setting.} For every pointer advance, the lag is the number of chunks from the advance to the milestone of the stage it leaves, negative when the milestone came first. An advance is grounded when this milestone arrives at most three chunks after it, and an episode is aligned when its final pointer lies within one stage of its completed stages. Panels (a, b) of Figure~\ref{fig:rq3_encoder} use the 500 test episodes, (c) the 200 episodes of Table~\ref{tab:main_results} (VLA-Adapter: 3 seeds), and (d) all 1,537 language variants by level.

\begin{figure}[!htb]
\centering
\includegraphics[width=\textwidth]{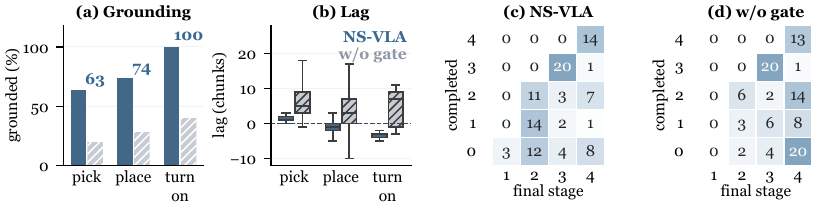}
\caption{Pointer on the 500 LIBERO-Long test episodes: (a) grounded advances and (b) lag per primitive left, with and without the gate; (c, d) final pointer stage against completed stages.}
\label{fig:app_pointer}
\end{figure}

\textbf{Results.} NS-VLA grounds $68.2\%$ of its 940 advances against $23.2\%$ of 1,220 without the gate, aligns $87.4\%$ of its episodes against $68.0\%$, and lowers the median lag from five chunks to one. The gain holds for every primitive the pointer leaves, $63$ against $20\%$ for \texttt{pick}, $74$ against $28\%$ for \texttt{place} and $100$ against $40\%$ for \texttt{turn on} (Figure~\ref{fig:app_pointer}), and the gate removes most runaway episodes that reach the last stage without completing any. NS-VLA completes $55.1\%$ of plan stages on the 200 episodes against $51.6\%$ for BAKU, and on paraphrases reaches $68.3$, $69.3$, $68.3$, $66.2$ and $32.2\%$ from difficulty level 1 to 5, against $64.3$, $61.8$, $48.6$, $46.0$ and $30.1\%$ for its VLA-Adapter backbone.

\subsection{Neuro-Symbolic Solver (RQ4)}
\textbf{Setting.} Figure~\ref{fig:rq4_solver} averages over training seeds 7, 17 and 27: (a) validation SR of every snapshot and (b) the L1 loss smoothed over 25 logged points. Panel (c) takes the 200 LIBERO-Long episodes of Table~\ref{tab:main_results} for all methods (VLA-Adapter: 3 seeds) and (d) the 500 test episodes of each seed.

\textbf{Results.} At 20k steps NS-VLA reaches $50.7\%$ mean validation SR against $34.0\%$, at loss $0.058$ against $0.078$. NS-VLA completes the task in $57.9\%$ of its 145 episodes that finish the first primitive, against $47.0\%$ of 117 for SmolVLA, and averages $45.3\pm9.6$ over seeds against $29.5\pm25.6$.

\subsection{Hierarchical Joint Policy Optimization (RQ5)}
\textbf{Setting.} Each iteration collects 80 rollouts (8 per task) from training states; the flat baseline runs the same stage on VLA-Adapter without plan or bridged input. Panel (c) of Figure~\ref{fig:rq5_rl} pairs the 500 test episodes of $\theta_0$ and the selected model, and (d) covers all 19 online runs: 3 flat on Long and 16 of NS-VLA, with the one-slot runs of Table~\ref{tab:main_results} and exploratory multi-slot and repeat runs only here.

\begin{table}[!htb]
\centering
\caption{Online stage on LIBERO-Long (seed 7): rollout SR (\%) and milestones per rollout.}
\label{tab:app_online}
\vspace{2pt}
\fontsize{8pt}{10pt}\selectfont
\setlength{\tabcolsep}{4.5pt}
\begin{tabular}{@{}llcccccccc@{}}
\toprule
& & \textbf{1} & \textbf{2} & \textbf{3} & \textbf{4} & \textbf{5} & \textbf{6} & \textbf{7} & \textbf{8} \\
\midrule
\multirow{2}{*}{Rollout SR} & flat RL & 0.0 & 0.0 & 0.0 & 0.0 & 0.0 & 0.0 & 0.0 & 0.0 \\
 & \cellcolor{nslight}\textbf{NS-VLA} & \cellcolor{nslight}\textbf{32.5} & \cellcolor{nslight}\textbf{22.5} & \cellcolor{nslight}\textbf{38.8} & \cellcolor{nslight}\textbf{28.8} & \cellcolor{nslight}\textbf{40.0} & \cellcolor{nslight}\textbf{36.3} & \cellcolor{nslight}\textbf{18.8} & \cellcolorR{nslight}\textbf{26.3} \\
\midrule
\multirow{2}{*}{Milestones} & flat RL & 0.25 & 0.19 & 0.16 & 0.23 & 0.18 & 0.24 & 0.31 & 0.25 \\
 & \cellcolor{nslight}\textbf{NS-VLA} & \cellcolor{nslight}\textbf{1.76} & \cellcolor{nslight}\textbf{1.53} & \cellcolor{nslight}\textbf{1.88} & \cellcolor{nslight}\textbf{1.64} & \cellcolor{nslight}\textbf{2.01} & \cellcolor{nslight}\textbf{1.68} & \cellcolor{nslight}\textbf{1.39} & \cellcolorR{nslight}\textbf{1.49} \\
\bottomrule
\end{tabular}
\end{table}

\textbf{Results.} NS-VLA collects successes in every iteration (Table~\ref{tab:app_online}) and reaches $1.67$ milestones per rollout, $7.4\times$ the $0.225$ of the flat baseline, whose replay set stays empty. On the test episodes the online stage turns 105 failures into successes, a net gain of 62. Selection improves on the supervised start in 9 of the 19 runs, by 4 to 8 validation points, and matches it in the other 10.

\subsection{Case study (RQ6)}
\begin{figure}[!htb]
\centering
\includegraphics[width=\textwidth]{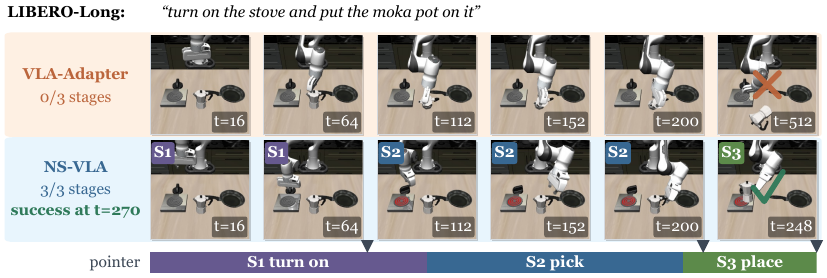}
\caption{A second case, whose plan begins with the grasp--release primitive \texttt{turn on}.}
\label{fig:app_case2}
\end{figure}

\textbf{Setting.} Rollouts use the reported checkpoints from official initial state 0: LIBERO-Long task 3 and its paraphrase (language variant 1225, level 4) in Figure~\ref{fig:rq6_case}, and task 2 in Figure~\ref{fig:app_case2}.

\textbf{Results.} NS-VLA turns on the stove, picks up the moka pot and places it on the stove, with milestones at steps 88, 224 and 270, and succeeds at step 270, while VLA-Adapter reaches no stage. The pointer leaves every stage, \texttt{turn on} included, only once the gate sees the required gripper evidence.

\subsection{Real-world tasks (RQ7)}
\textbf{Setting.} The myCobot 320pi arm observes the scene with a wrist-mounted and a third-person RealSense D435i camera (Figure~\ref{fig:rq7_real}). Three Pick, three Move and four Place tasks are each repeated ten times under randomized object placements for both policies (30, 30 and 40 trials).

\textbf{Results.} NS-VLA succeeds in 22 of 30 Pick, 18 of 30 Move and 20 of 40 Place, against 16, 12 and 12 for VLA-Adapter, leading each category by 20 points and reaching $60.0$ against $40.0\%$ overall.

\section{Extended related work}
\label{app:ext_related}

\textbf{Reinforcement learning for robot policies.} Trust-region and clipped policy gradients with advantage estimation~\citep{Trust-Region-Policy-Optimization, Proximal-Policy-Optimization-Algorithms, GAE} are the standard objectives of online RL, while advantage-weighted and offline methods keep the policy close to the data~\citep{Advantage-Weighted-Regression, AWAC, Behavior-Regularized-Offline-Reinforcement-Learning, A-Minimalist-Approach-to-Offline-Reinforcement-Learning, Behavior}. Preference and group-relative objectives from language modeling~\citep{Training-language-models-to-follow-instructions-with-human-feedback, DPO, DeepSeekMath} were carried to VLAs and embodied reasoners~\citep{VLA-RL, VLA-R1, SimpleVLA-RL, VLA-RFT, SRPO, ManipLVM-R1, EVOLVE-VLA, Self-Evolved-Imitation-Learning}, and reward shaping densifies sparse returns~\citep{ng1999policy, wiewiora2003potential}. HJPO keeps the binary reward unshaped and makes it reachable through structure, turning every success into a parameter-free regression target for the action policy, with no critic and no learned reward model.

\textbf{Structure in language-conditioned manipulation.} Language models decompose instructions into skills, code or value maps~\citep{Do-As-I-Can-Not-As-I-Say, Code-as-Policies, Inner-Monologue, VoxPoser}, and VLAs reason through language hierarchies, chains of thought, visual traces or predicted futures~\citep{RT-H, ECoT, CoT-VLA, tracevla, Predictive-Inverse-Dynamics-Models-are-Scalable-Learners-for-Robotic-Manipulation}. Options, feudal managers and goal-conditioned hierarchies organize control in time~\citep{Between-MDPs-and-semi-MDPs, The-option-critic-architecture, Feudal-Reinforcement-Learning, vezhnevets2017feudal, Data-Efficient-Hierarchical-Reinforcement-Learning, Hierarchical-Deep-Reinforcement-Learning, Learning-Multi-Level-Hierarchies-with-Hindsight}, and neuro-symbolic planners execute parsed programs~\citep{NeSyPr, pmlr}. NS-VLA joins these lines within one chunked VLA policy.

\textbf{Hierarchical VLAs and their interfaces.} Two-level control is the setting of NS-VLA, not its claim; hierarchical VLAs differ in what crosses the interface (Table~\ref{tab:app_hier}). LCB~\citep{LCB} lets a language model emit a latent code that steers a pretrained policy, VAMOS~\citep{VAMOS} separates a general planner from an embodiment-specific capability model, and VINE~\citep{VINE} derives its hierarchy from success and failure demonstrations. NS-VLA passes a typed primitive, its free-text arguments and an observable pointer state, so the active primitive and every transition can be read off, certified and measured. A capability model decides which plans are admissible, this interface how a plan is executed and verified.

\begin{table}[!htb]
\centering
\caption{What crosses the interface between the two levels of recent hierarchical VLAs.}
\label{tab:app_hier}
\vspace{2pt}
\fontsize{8pt}{10pt}\selectfont
\setlength{\tabcolsep}{4pt}
\begin{tabular}{@{}l>{\raggedright\arraybackslash}p{0.38\linewidth}>{\raggedright\arraybackslash}p{0.38\linewidth}@{}}
\toprule
\textbf{Method} & \textbf{Interface between the levels} & \textbf{Source of the hierarchy} \\
\midrule
LCB~\citep{LCB} & continuous latent code & language model fine-tuned to emit the code \\
VAMOS~\citep{VAMOS} & navigation plan admitted by a capability model & general planner and embodiment-specific model \\
VINE~\citep{VINE} & plan scored by a success estimator & success and failure demonstrations \\
\rowcolor{nslight}
\shadeL\textbf{NS-VLA} & typed primitive, free-text arguments and pointer state & compositional structure of the instruction\shadeR \\
\bottomrule
\end{tabular}
\end{table}

\section{Scaling the symbolic layer}
\label{app:scaling}

\textbf{Division of labor.} The symbolic layer carries only the plan structure and the execution gate; perception, grounding and control stay neural. Operations form a small vocabulary, while objects, supports and relations stay free-text arguments, so the unbounded side of the world needs no symbol. The pointer reads its features from the fine-tuned backbone, so no frozen encoder bounds the visual domain. Unlike classical expert systems, whose rules carried perception and control and failed hard at every gap, a gap in the grammar costs only the decomposition: an instruction that is not decomposed runs with the one-slot plan, on which NS-VLA is exactly its backbone (Proposition~\ref{prop:graceful_equiv}).

\textbf{What a new task costs.} Only a new operation requires retraining (Table~\ref{tab:app_cost}), and it costs a grammar rule and one pass of the automatic pipeline rather than a manual annotation campaign.

\begin{table}[!htb]
\centering
\caption{What NS-VLA retrains, and what it keeps, when a new task brings something new.}
\label{tab:app_cost}
\vspace{2pt}
\fontsize{8pt}{10pt}\selectfont
\setlength{\tabcolsep}{4pt}
\begin{tabular}{@{}>{\raggedright\arraybackslash}p{0.36\linewidth}>{\raggedright\arraybackslash}p{0.58\linewidth}@{}}
\toprule
\textbf{What is new} & \textbf{What is retrained} \\
\midrule
an object, support or relation & nothing: it enters as a free-text argument \\
a phrasing that the grammar parses & nothing: the same plan is produced \\
a phrasing that the grammar does not parse & nothing: the one-slot plan runs the backbone exactly \\
an operation & a grammar rule with its gripper class, automatic segmentation of its demonstrations, and the supervised stage on them \\
\bottomrule
\end{tabular}
\end{table}

\textbf{Adaptation within an episode.} The plan depends on the instruction alone, so recompiling it mid-episode returns the same plan; the pointer decides from observations when a primitive is done, and the Solver recovers inside a primitive, e.g.\ by regrasping while the gate holds the pointer on \texttt{pick}.

\section{Limitations and future work}
\label{sec:future_directions}

\textbf{Limitations.} The operations come from a grammar: a new operation needs a rule and its gripper class, and an instruction the grammar does not decompose runs as the plain backbone, without plan or pointer. The plan is fixed per episode and the pointer only stays or advances, so a stage left too early is not revisited and the plan content is not revised when the scene contradicts it. The gate certifies grasp and release primitives from proprioception, whereas gripper-free primitives rely on the classifier and the dwell. The evaluation covers the four LIBERO suites with five demonstrations per task, LIBERO-Plus and one real arm with 100 trials per policy, with VLA-Adapter as the only backbone and seed 7 for the main table; online training needs simulated rollouts with binary success.

\textbf{Future work.} Four directions build on the interface: a pointer that may also retreat to an earlier stage; observation-triggered revision of the plan, such as inserting ``open the drawer'' when the drawer is found shut; label-free discovery of operations by clustering contact and gripper events across demonstrations; and object-centric arguments, where single-image 3D reconstruction such as SAM 3D~\citep{SAM3D} supplies per-object shape and pose, so that a novel object needs no new symbol and grasping and placing become directly observable. Since the interface needs only an instruction-conditioned chunked policy, transferring it to further VLA backbones and robot platforms is a direct next step.

\section{Broader impact}
\label{sec:broader_impact}

Learning multi-step manipulation from few demonstrations lowers the cost of teaching robots household and assistive tasks, and the explicit plan makes every decision readable to an operator. As NS-VLA reduces exactly to its backbone on single-primitive instructions and selection never falls below the supervised start on validation, adopting it preserves the behavior of an existing policy wherever structure is absent. A misparsed instruction or a failed grasp can still move objects unexpectedly, so deployment should follow standard robot-safety practice, with the observable stage pointer as a natural point for monitoring and human intervention at every plan stage.


\newpage

\section*{NeurIPS Paper Checklist}

\begin{enumerate}

\item {\bf Claims}
    \item[] Question: Do the main claims made in the abstract and introduction accurately reflect the paper's contributions and scope?
    \item[] Answer: \answerYes{}
    \item[] Justification: The abstract and \S\ref{Section1} state the three components of NS-VLA (the Neuro-Symbolic Encoder, the Neuro-Symbolic Solver and Hierarchical Joint Policy Optimization) and the setting: few demonstrations (five per task, 10\% of LIBERO), held-out LIBERO-Plus evaluation and a real robot arm. Each component is backed by a proposition (Propositions~\ref{prop:ns_consistency}, \ref{prop:graceful_equiv}, \ref{prop:rl_convergence}) and by the experiments of \S\ref{sec:experiments}: the comparison with ten baselines and VLA-Adapter (Table~\ref{tab:main_results}), the ablation (Table~\ref{tab:ablation}), the component analyses (RQ3--RQ5), the case study (RQ6) and the real-robot study (RQ7).

\item {\bf Limitations}
    \item[] Question: Does the paper discuss the limitations of the work performed by the authors?
    \item[] Answer: \answerYes{}
    \item[] Justification: Appendix~\ref{sec:future_directions} (Limitations and future work) discusses the limitations: the grammar-defined operations and the one-slot fallback, the fixed plan and the monotone pointer, the gripper-based gate, and the evaluation scope (four LIBERO suites with five demonstrations per task, LIBERO-Plus, one real arm, VLA-Adapter as the backbone). The assumptions of each proposition are stated with it, and the evaluation protocol is specified in \S\ref{sec:setup} and Appendix~\ref{sec:dataset_details}.

\item {\bf Theory assumptions and proofs}
    \item[] Question: For each theoretical result, does the paper provide the full set of assumptions and a complete (and correct) proof?
    \item[] Answer: \answerYes{}
    \item[] Justification: The paper contains three propositions, each stated with its assumptions in \S\ref{sec:method}: the stage pointer splits every episode into at most $M$ ordered contiguous segments, each non-final one lasting at least $d_{\min}$ decisions and left only once the gate certifies it (Proposition~\ref{prop:ns_consistency}); with a single-primitive plan NS-VLA reduces exactly to its backbone (Proposition~\ref{prop:graceful_equiv}); and filtered advantage-weighted regression is the $\beta\to 0$ limit of advantage-weighted regression under binary task reward, with validation-based selection never below the supervised start (Proposition~\ref{prop:rl_convergence}). Complete proofs are given in Appendix~\ref{proof1}, Appendix~\ref{proof2} and Appendix~\ref{proof3}, respectively.

    \item {\bf Experimental result reproducibility}
    \item[] Question: Does the paper fully disclose all the information needed to reproduce the main experimental results of the paper to the extent that it affects the main claims and/or conclusions of the paper (regardless of whether the code and data are provided or not)?
    \item[] Answer: \answerYes{}
    \item[] Justification: The architecture, the deterministic plan parser and segmentation rule, the pointer and gate thresholds, the supervised and online training schedules, the selection rule, the 10\% demonstration split, the LIBERO-Plus sampling (seed 2026 and the pre-registered extension with seed 2027) and the evaluation protocol (seed 7, official initial states, episode limits) are documented in \S\ref{sec:setup} and Appendices~\ref{sec:rl_details}, \ref{sec:baseline_details}, \ref{sec:dataset_details} and \ref{sec:implementation_details} (Table~\ref{tab:hyperparams}); the upstream code version of every baseline is given in Appendix~\ref{sec:baseline_details} and its training configuration in Table~\ref{tab:baseline_budget}. Public code, model checkpoints and dataset resources accompany the paper.

\item {\bf Open access to data and code}
    \item[] Question: Does the paper provide open access to the data and code, with sufficient instructions to faithfully reproduce the main experimental results, as described in supplemental material?
    \item[] Answer: \answerYes{}
    \item[] Justification: The \href{https://github.com/Zuzuzzy/NS-VLA}{public code repository} provides the training, online-stage and evaluation scripts with their configuration files; the \href{https://huggingface.co/zuzuzzy/NS-VLA}{model checkpoints} and \href{https://huggingface.co/datasets/zuzuzzy/NS-VLA-Dataset}{released dataset} are publicly available; LIBERO and LIBERO-Plus are accessed through their official releases, and the baseline codebases at the commits and versions listed in Appendix~\ref{sec:baseline_details}.

\item {\bf Experimental setting/details}
    \item[] Question: Does the paper specify all the training and test details (e.g., data splits, hyperparameters, how they were chosen, type of optimizer) necessary to understand the results?
    \item[] Answer: \answerYes{}
    \item[] Justification: The training split (first five demonstrations per task), the held-out LIBERO-Plus evaluation, the optimizer, learning rates, batch sizes, step counts, LoRA rank, chunk length, sub-instruction probability, pointer thresholds, online-stage schedule (iterations, rollouts, exploration noise, demonstration ratio) and the validation-based checkpoint and iterate selection are specified in \S\ref{sec:setup} and Appendix~\ref{sec:implementation_details}; the baseline configurations are given in Appendix~\ref{sec:baseline_details}.

\item {\bf Experiment statistical significance}
    \item[] Question: Does the paper report error bars suitably and correctly defined or other appropriate information about the statistical significance of the experiments?
    \item[] Answer: \answerYes{}
    \item[] Justification: Table~\ref{tab:main_results} reports, for every suite and dimension, the binomial standard error over evaluation episodes, $\sqrt{p(1-p)/n}$, and for the averages the standard error propagated from these; a cell with no success shows the one-sided 95\% Clopper--Pearson interval $[0,u]$ as $u/2 \pm u/2$, so no bar leaves $[0,100]$, and the LIBERO-Long cell of VLA-Adapter is the mean over three training seeds with the standard error across seeds. The LIBERO-Plus comparison with VLA-Adapter is tested by an exact McNemar test over the 1,200 paired variants. The ablation reports the SR change of every variant with a 95\% interval from a task-clustered paired bootstrap over the ten tasks (10,000 resamples), which keeps episodes paired across variants and makes no normality assumption (Table~\ref{tab:app_ablation}). The Solver analysis (Figure~\ref{fig:rq4_solver}) reports the mean and sample standard deviation of test SR over three training seeds and the range over seeds of the validation curve. All comparisons use identical episodes for every method (\S\ref{sec:setup}, Appendix~\ref{sec:dataset_details}).

\item {\bf Experiments compute resources}
    \item[] Question: For each experiment, does the paper provide sufficient information on the computer resources (type of compute workers, memory, time of execution) needed to reproduce the experiments?
    \item[] Answer: \answerYes{}
    \item[] Justification: NS-VLA and VLA-Adapter are trained on a single NVIDIA H100 GPU per run in BF16. For NS-VLA, the supervised stage at batch 16 takes 2.6--4.6\,h for 10k steps (Spatial, Object, Goal) and 5.2--5.4\,h for 20k steps (Long) with a peak memory below 39\,GB, and the online stage (8 iterations of 80 rollouts and 400 updates) takes 1.5--3.1\,h. The ten baselines are trained on NVIDIA V100 (32\,GB) GPUs with the per-suite configurations of Table~\ref{tab:baseline_budget}, and all methods are evaluated on the same machines with EGL rendering (Appendices~\ref{sec:baseline_details} and \ref{sec:implementation_details}). Preliminary and diagnostic runs that are not reported in the paper used additional compute of the same type.
    
\item {\bf Code of ethics}
    \item[] Question: Does the research conducted in the paper conform, in every respect, with the NeurIPS Code of Ethics \url{https://neurips.cc/public/EthicsGuidelines}?
    \item[] Answer: \answerYes{}
    \item[] Justification: The research uses publicly available simulation benchmarks and a tabletop robot arm in a laboratory setting, involves no human subjects or personal data, and conforms to the NeurIPS Code of Ethics in all respects.

\item {\bf Broader impacts}
    \item[] Question: Does the paper discuss both potential positive societal impacts and negative societal impacts of the work performed?
    \item[] Answer: \answerYes{}
    \item[] Justification: Appendix~\ref{sec:broader_impact} discusses the societal impact. Learning multi-step manipulation from five demonstrations per task lowers the cost of teaching robots household and assistive tasks, and the explicit plan makes every decision readable to an operator. As a negative impact, a misparsed instruction or a failed grasp can move objects unexpectedly; deployment should therefore follow standard robot-safety practice, with the observable stage pointer as a point for monitoring and human intervention, and NS-VLA reduces exactly to its backbone where an instruction has no multi-step structure.
    
\item {\bf Safeguards}
    \item[] Question: Does the paper describe safeguards that have been put in place for responsible release of data or models that have a high risk for misuse (e.g., pre-trained language models, image generators, or scraped datasets)?
    \item[] Answer: \answerNA{}
    \item[] Justification: The released artifact is a robotic manipulation policy fine-tuned on publicly available simulation benchmarks; it does not constitute a generative model, a scraped dataset, or any artifact that poses a high risk for misuse.

\item {\bf Licenses for existing assets}
    \item[] Question: Are the creators or original owners of assets (e.g., code, data, models), used in the paper, properly credited and are the license and terms of use explicitly mentioned and properly respected?
    \item[] Answer: \answerYes{}
    \item[] Justification: All assets are publicly released by their original authors, cited in the paper and used under their terms: the LIBERO benchmark and demonstrations~\citep{LIBERO} (MIT), the LIBERO-Plus benchmark~\citep{LIBERO-Plus} (official release; no license file is declared in its repository), the VLA-Adapter backbone~\citep{VLA-Adapter} (MIT) with its Qwen2.5-0.5B (Apache-2.0), DINOv2 (Apache-2.0) and SigLIP (Apache-2.0) components, and the baseline codebases and pretrained weights: LIBERO (MIT), QueST (MIT), BAKU (MIT), Octo and octo-base-1.5 (MIT), and LeRobot with smolvla\_base (Apache-2.0), with the pretrained BERT, T5 and MiniLM (Apache-2.0), CLIP (MIT) and SmolVLM2 (Apache-2.0) encoders they use, whose exact versions are listed in Appendix~\ref{sec:baseline_details}.

\item {\bf New assets}
    \item[] Question: Are new assets introduced in the paper well documented and is the documentation provided alongside the assets?
    \item[] Answer: \answerYes{}
    \item[] Justification: The \href{https://github.com/Zuzuzzy/NS-VLA}{public code repository} is released under the Apache-2.0 license and includes a README with installation instructions, training and evaluation commands, and configuration files. The \href{https://huggingface.co/zuzuzzy/NS-VLA}{model checkpoints} and \href{https://huggingface.co/datasets/zuzuzzy/NS-VLA-Dataset}{released dataset} are documented on their respective artifact pages.

\item {\bf Crowdsourcing and research with human subjects}
    \item[] Question: For crowdsourcing experiments and research with human subjects, does the paper include the full text of instructions given to participants and screenshots, if applicable, as well as details about compensation (if any)?
    \item[] Answer: \answerNA{}
    \item[] Justification: The paper does not involve crowdsourcing or research with human subjects; all experiments are conducted in simulation or with a robot arm operated by the authors.

\item {\bf Institutional review board (IRB) approvals or equivalent for research with human subjects}
    \item[] Question: Does the paper describe potential risks incurred by study participants, whether such risks were disclosed to the subjects, and whether Institutional Review Board (IRB) approvals (or an equivalent approval/review based on the requirements of your country or institution) were obtained?
    \item[] Answer: \answerNA{}
    \item[] Justification: The paper does not involve research with human subjects; no IRB approval or equivalent is required.

\item {\bf Declaration of LLM usage}
    \item[] Question: Does the paper describe the usage of LLMs if it is an important, original, or non-standard component of the core methods in this research? Note that if the LLM is used only for writing, editing, or formatting purposes and does \emph{not} impact the core methodology, scientific rigor, or originality of the research, declaration is not required.
    \item[] Answer: \answerYes{}
    \item[] Justification: NS-VLA builds on VLA-Adapter, whose Qwen2.5-0.5B language model, adapted with LoRA rank 64, is a core component of the policy and is described in \S\ref{sec:setup} and Appendix~\ref{sec:implementation_details}. The primitive plan is produced by the deterministic grammar parser of \S\ref{sec:ns_encoder}, so no language-model prompting is involved in plan extraction.

\end{enumerate}

\end{document}